\documentclass[10pt,twocolumn,letterpaper]{article}

\usepackage{ieee}
\usepackage{times}
\usepackage{epsfig}
\usepackage{graphicx}
\usepackage{amsmath}
\usepackage{amssymb}
\usepackage{scalerel}
\usepackage{siunitx}
\usepackage[nocompress]{cite}
\usepackage{enumitem}
\usepackage{algpseudocode}
\usepackage{algorithm}
\usepackage{multirow}
\usepackage[title]{appendix}
\usepackage{subfigure} % [FIGTOPCAP]

\usepackage[pagebackref=true,breaklinks=true,letterpaper=true,colorlinks,bookmarks=false]{hyperref}

\usepackage{xcolor}
\definecolor{colorall}{rgb}{0 0.60 0.07}
\definecolor{colorlink}{rgb}{0.80 0 0.07}
\definecolor{colorfile}{rgb}{0 0.60 0.07}
\definecolor{colorurl}{HTML}{E8688E}
\hypersetup{
    colorlinks=true,
    allcolors=colorall,
    linkcolor=colorlink,
    filecolor=colorfile,      
    urlcolor=colorurl,
}

\usepackage{cleveref}

\ieeefinalcopy
\def\ieeePaperID{-}

\newcommand{\ourmethod}{\mbox{EfficientAD}}
\newcommand{\patchcoreens}{PatchCore\textsubscript{\scaleto{\text{Ens}}{5pt}}}
\newcommand{\studteach}{\mbox{S--T}}
\newcommand{\unet}{\mbox{U-Net}}
\newcommand{\meanwithstd}[2]{\specialcell[c]{#1 \\[-3pt] {\footnotesize ($\pm$ #2)}}}
\newcommand\Tstrut{\rule{0pt}{2.6ex}}
\newcommand\Bstrut{\rule[-0.9ex]{0pt}{0pt}}
\newcommand{\specialcell}[2][c]{%
  \begin{tabular}[#1]{@{}c@{}}#2\end{tabular}}

\begin{document}

\title{\ourmethod: Accurate Visual Anomaly Detection at Millisecond-Level Latencies}

\author{Kilian Batzner \hspace{1cm} Lars Heckler \hspace{1cm} Rebecca König \vspace{0.14cm} \\
MVTec Software GmbH\\
{\tt\small \{kilian.batzner, lars.heckler, rebecca.koenig\}@mvtec.com}
}

\maketitle

\begin{abstract}

Detecting anomalies in images is an important task, especially in real-time computer vision applications.
In this work, we focus on computational efficiency and propose a lightweight feature extractor that processes an image in less than a millisecond on a modern GPU.
We then use a student--teacher approach to detect anomalous features.
We train a student network to predict the extracted features of normal, i.e., anomaly-free training images.
The detection of anomalies at test time is enabled by the student failing to predict their features.
We propose a training loss that hinders the student from imitating the teacher feature extractor beyond the normal images.
It allows us to drastically reduce the computational cost of the student--teacher model, while improving the detection of anomalous features.
We furthermore address the detection of challenging logical anomalies that involve invalid combinations of normal local features, for example, a wrong ordering of objects.
We detect these anomalies by efficiently incorporating an autoencoder that analyzes images globally.
We evaluate our method, called \ourmethod, on 32 datasets from three industrial anomaly detection dataset collections.
\ourmethod\ sets new standards for both the detection and the localization of anomalies.
At a latency of two milliseconds and a throughput of six hundred images per second, it enables a fast handling of anomalies.
Together with its low error rate, this makes it an economical solution for real-world applications and a fruitful basis for future research.

\end{abstract}

\section{Introduction}

In the past years, deep learning methods have continued to improve the state of the art across a wide range of computer vision applications.
This progress has been accompanied by advances in making neural network architectures faster and more efficient \cite{tan2019efficientnet, tan2020efficientdet, redmon2016you, yin2022vit}.
Modern classification architectures, for example, focus on characteristics such as latency, throughput, memory consumption, and the number of trainable parameters \cite{tan2019efficientnet, tan2021efficientnetv2, liu2022convnet, yin2022vit, sandler2018mobilenetv2, liu2021swin}.
This ensures that as networks become more capable, their computational requirements remain suitable for real-world applications. 
The field of visual anomaly detection has also seen rapid progress in the recent past, especially on industrial anomaly detection benchmarks \cite{bergmann2021_mvtec_ad_ijcv, bergmann2019_mvtec_ad_cvpr, roth2022towards, rudolph2023asymmetric}.
State-of-the-art anomaly detection methods, however, often sacrifice computational efficiency for an increased anomaly detection performance.
Common techniques are ensembling, the use of large backbones, and increasing the input image resolution to up to 768$\times$768 pixels.

\begin{figure}[t]
\begin{center}
\includegraphics[width=1.0\linewidth]{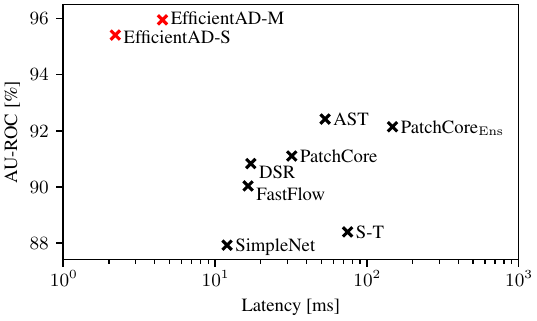}
\end{center}
   \caption{Anomaly detection performance vs. latency per image on an NVIDIA RTX A6000 GPU\@. Each AU-ROC value is an average of the image-level detection AU-ROC values on the MVTec AD \cite{bergmann2021_mvtec_ad_ijcv, bergmann2019_mvtec_ad_cvpr}, VisA \cite{zou2022spot}, and MVTec LOCO \cite{bergmann2021_mvtec_loco_ijcv} dataset collections.}
\label{fig:teaser}
\end{figure}

Real-world anomaly detection applications frequently put constraints on the computational requirements of a method.
There are cases where detecting an anomaly too late can cause substantial economic damage, such as metal objects in a crop field entering the interior of a combine harvester.
In other cases, even human health is at risk, for example, if a limb of a machine operator approaches a blade.
Furthermore, industrial settings commonly involve strict runtime limits caused by high production rates \cite{bailey2012machine_vision_handbook_fpgas}.
Not adhering to these limits would decrease the production rate of the respective application and thus its economic viability.
It is therefore essential to pay attention to the computational and economic cost of anomaly detection methods to keep them suitable for real-world applications.

In this work, we propose \ourmethod, a method that sets new standards for both the anomaly detection performance and the inference runtime, as shown in \Cref{fig:teaser}.
We first introduce an efficient network architecture for computing expressive features in less than a millisecond on a modern GPU\@.
To detect anomalous features, we use a student--teacher approach \cite{bergmann2020_uninformed_cvpr, rudolph2023asymmetric, wang2021student_teacher}.
We train a student network to predict the features computed by a pretrained teacher network on normal, i.e., anomaly-free training images.
Because the student is not trained on anomalous images, it generally fails to mimic the teacher on these.
A large distance between the outputs of the teacher and the student thus enables the detection of anomalies at test time.
To further increase this effect, Rudolph \etal \cite{rudolph2023asymmetric} use \textit{architectural} asymmetry between the teacher and the student.
We instead propose \textit{loss-induced} asymmetry in the form of a training loss that hinders the student from imitating the teacher beyond the normal images.
This loss does not affect the computational cost at test time and does not restrict the architecture design.
It allows us to use our efficient network architecture for both the student and the teacher, while improving the detection of anomalous features.

Identifying anomalous local features enables the detection of anomalies that are \textit{structurally} different from the normal images, for example, contaminations or stains on manufactured products.
A challenging problem, however, are violations of \textit{logical} constraints regarding the position, size, arrangement, etc.\ of normal objects.
To address this, \ourmethod\ includes an autoencoder that learns the logical constraints of training images and detects violations at test time.
We show how to integrate the autoencoder efficiently with a student--teacher model.
Furthermore, we present a method to improve the anomaly detection performance by calibrating the detection results of the autoencoder and the student--teacher model before combining their results.

Our contributions are summarized as follows:
\begin{itemize}[topsep=0.2em]
    \itemsep-0.2em
    \item We substantially improve the state of the art for both the detection and the localization of anomalies on industrial benchmarks, at a latency of \SI{2}{\milli\second} and a throughput of more than 600 images per second.
    \item We propose an efficient network architecture to speed up feature extraction by an order of magnitude in comparison to the feature extractors used by recent methods \cite{roth2022towards, rudolph2023asymmetric, yu2021fastflow}.
    \item We introduce a training loss that significantly improves the anomaly detection performance of a student--teacher model without affecting its inference runtime.
    \item We achieve an efficient autoencoder-based detection of logical anomalies and propose a method for a calibrated combination of the detection results with those of a student--teacher model.
\end{itemize}

\section{Related Work}

\subsection{Anomaly Detection Tasks}

Visual anomaly detection is a rapidly growing area of research with a diverse range of applications, including medical imaging \cite{fernando2021_medical_ad_survey, armato2011lung, menze2015_brats_dataset}, autonomous driving \cite{hendrycks2019scaling, lis2019_iccv_resynthesis, blum2019_fishyscapes_dataset}, and industrial inspection \cite{bergmann2021_mvtec_ad_ijcv, ehret2019_ad_review_paper, pang2020_review_paper}.
Applications often have specific characteristics, such as the availability of image sequences in surveillance datasets \cite{li2013_ucsd_video_ad_dataset, zhang2016single, lu2013_avenue_video_ad_dataset} or the different modalities of medical imaging datasets (MRI \cite{bakas2017_brats_dataset}, CT \cite{armato2011lung}, X-ray \cite{irvin2019chexpert}, etc.).
This work focuses on detecting anomalies in RGB or gray-scale images without conditioning the prediction on a sequence of images.
We use industrial anomaly detection datasets to benchmark our proposed method against existing ones.

The introduction of the MVTec AD dataset \cite{bergmann2019_mvtec_ad_cvpr, bergmann2021_mvtec_ad_ijcv} has catalyzed the development of methods for industrial applications.
It comprises 15 separate inspection scenarios, each consisting of a training set and a test set.
Each training set contains only normal images, for example, defect-free screws, while the test sets also contain anomalous images.
This represents a frequent challenge in real-world applications where the types and possible locations of defects are unknown during the development of the anomaly detection system.
Therefore, it is a challenging yet crucial requirement that methods perform well when trained only on normal images.

Recently, several new industrial anomaly detection datasets have been introduced \cite{bergmann2022_mvtec_3dad, bergmann2021_mvtec_loco_ijcv, zou2022spot, mishra2021vt, huang2018_magnetic_tile_dataset, jezek2021deep}.
The Visual Anomaly (VisA) dataset \cite{zou2022spot} and the MVTec Logical Constraints (MVTec LOCO) dataset \cite{bergmann2021_mvtec_loco_ijcv} follow the design of MVTec AD and comprise twelve and five anomaly detection scenarios, respectively.
They contain anomalies that are empirically more challenging than those of MVTec AD\@.
Furthermore, MVTec LOCO contains not only structural anomalies, such as stains or scratches, but also logical anomalies.
These are violations of logical constraints, for example, a wrong ordering or a wrong combination of normal objects.
We refer to MVTec AD, VisA, and MVTec LOCO as dataset collections, as each scenario is a separate dataset consisting of a training and a test set.
All three provide pixel-precise defect segmentation masks for evaluating the anomaly localization performance of a method.

\subsection{Anomaly Detection Methods}

Traditional computer vision algorithms have been applied successfully to industrial anomaly detection tasks for several decades \cite{steger2018_mva_book}.
These algorithms commonly fulfill the requirement of processing an image within a few milliseconds.
Bergmann \etal \cite{bergmann2021_mvtec_ad_ijcv} evaluate some of these methods and find that they fail when requirements such as well-aligned objects are not met.
Deep-learning-based methods have been shown to handle such cases more robustly \cite{bergmann2021_mvtec_ad_ijcv, bergmann2021_mvtec_loco_ijcv}.

A successful approach in the recent past has been to apply outlier detection and density estimation methods in the feature space of a pretrained and frozen convolutional neural network (CNN).
If feature vectors can be mapped to input pixels, assigning their outlier scores to the respective pixels yields a 2D anomaly map of pixel anomaly scores.
Common methods include multivariate Gaussian distributions \cite{defard2021_PaDiM, rippel2021_Gaussian, li2021cutpaste}, Gaussian Mixture Models \cite{mishra2021vt, zong2018deep}, Normalizing Flows \cite{yu2021fastflow, rudolph2022_cross_flows, gudovskiy2022_CFLOW, rudolph2021_differnet, rezende2015variational}, and the k-Nearest Neighbor (kNN) algorithm \cite{napoletano2018_cnn_feature_dictionary_nanofibres, cohen2020_subimage, roth2022towards, nazare2018_pretrained_cnns_for_ad}.
A runtime bottleneck for kNN-based methods is the search for nearest neighbors during inference.
With PatchCore \cite{roth2022towards}, Roth \etal therefore perform kNN on a reduced database of clustered feature vectors.
They achieve state-of-the-art anomaly detection results on MVTec AD\@.
In our experiments, we include PatchCore and FastFlow \cite{yu2021fastflow}, a recent Normalizing-Flow-based method with a comparatively low inference runtime.

Bergmann \etal \cite{bergmann2020_uninformed_cvpr} propose a student--teacher (\studteach) framework for anomaly detection, in which the teacher is a pretrained frozen CNN\@.
They train student networks to mimic the output of the teacher on the training images.
Because the students have not seen anomalous images during training, they generally fail to predict the teacher's output on these images, which enables anomaly detection.
Various modifications of \studteach\ have been proposed \cite{salehi2021_st_ad, wang2021student_teacher, rudolph2023asymmetric}.
Rudolph \etal \cite{rudolph2023asymmetric} reach a competitive anomaly detection performance on MVTec AD by restricting the teacher to be an invertible neural network.
We compare our method to their Asymmetric Student Teacher (AST) approach and to the original \studteach\ method \cite{bergmann2020_uninformed_cvpr}.

Generative models such as autoencoders \cite{bergmann2018_ssim_ae, park2020_mmnad, baur2019_ano_vae_gan, liu2020_visually_explaining_vaes, sakurada2014_aes_for_ad, bergmann2021_mvtec_loco_ijcv, gong2019_mem_ae_iccv} and GANs \cite{goodfellow2014_gans, schlegl2017_anogan, schlegl2019_fast_anogan, perera2019_cvpr_ocgan, akcay2019ganomaly} have been used extensively for anomaly detection.
Recent autoencoder-based methods rely on accurate reconstructions of normal images and inaccurate reconstructions of anomalous images \cite{bergmann2018_ssim_ae, park2020_mmnad, bergmann2021_mvtec_loco_ijcv, gong2019_mem_ae_iccv}.
This enables detecting anomalies by comparing the reconstruction to the input image.
A common problem are false-positive detections caused by inaccurate reconstructions of normal images, e.g., blurry reconstructions.
To avoid this, GCAD \cite{bergmann2021_mvtec_loco_ijcv} lets an autoencoder reconstruct images in the feature space of a pretrained network.
Another recent reconstruction-based method is DSR \cite{zavrtanik2022dsr}, which uses the latent space of a pretrained autoencoder and generates synthetic anomalies in it.
Similarly, the recently proposed SimpleNet \cite{Liu_2023_CVPR} generates synthetic anomalies in a pretrained feature space to train a discriminator network for detecting anomalous features.
In our experiments, we include GCAD, DSR, and SimpleNet.

\section{Method}

We describe the components of \ourmethod\ in the following subsections.
It begins with the efficient extraction of features from a pretrained neural network in \cref{sec:method_pdn}.
We detect anomalous features at test time using a lightweight student--teacher model, as described in \cref{sec:method_structural}.
A key challenge is to achieve a competitive anomaly detection performance while keeping the overall runtime low.
To this end, we introduce a novel loss function for the training of a student--teacher model.
In \cref{sec:method_logical}, we explain how to efficiently detect logical anomalies with an autoencoder-based approach.
Finally, we provide a solution for calibrating and combining the detection results of the autoencoder with those of the student--teacher model in \cref{sec:method_balancing}.

\subsection{Efficient Patch Descriptors}
\label{sec:method_pdn}

\begin{figure}[t]
\begin{center}
\includegraphics[width=1.0\linewidth]{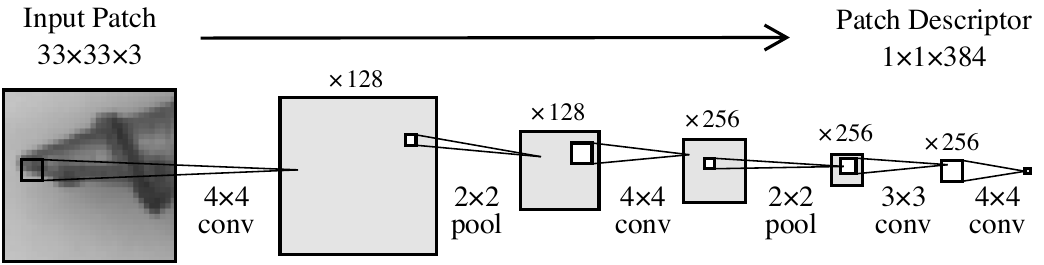}
\end{center}
   \caption{Patch description network (PDN) architecture of \ourmethod-S.
   Applying it to an image in a fully convolutional manner yields all features in a single forward pass.
   }
\label{fig:pdn}
\vspace*{2mm}
\end{figure}

Recent anomaly detection methods commonly use the features of a deep pretrained network, such as a WideResNet-101 \cite{zagoruyko2016wideresnet_wrn, roth2022towards}.
We use a network with a drastically reduced depth as a feature extractor.
It consists of only four convolutional layers and is visualized in \Cref{fig:pdn}.
Each output neuron has a receptive field of 33$\times$33 pixels and thus each output feature vector describes a 33$\times$33 patch.
Due to this clear correspondence, we refer to the network as a patch description network (PDN).
The PDN is fully convolutional and can be applied to an image of variable size to generate all feature vectors in a single forward pass.

The \studteach\ method \cite{bergmann2020_uninformed_cvpr} also uses features from networks with only few convolutional layers.
The computational cost of these networks is nevertheless high because of the lack of downsampling in convolutional and pooling layers.
The number of parameters of the networks used by \studteach\ is comparably low (between 1.6 and 2.7 million per network).
Yet, executing a single network takes longer and requires more memory in our experiments than a \unet\ \cite{ronneberger2015_u_net} with 31 million parameters, an architecture used by the GCAD method \cite{bergmann2021_mvtec_loco_ijcv}.
This demonstrates how the number of parameters can be a misleading proxy metric for the latency, throughput, and memory footprint of a method.
Modern classification architectures typically perform downsampling early to reduce the size of feature maps and thus the runtime and memory requirements \cite{he2016_resnet_paper}.
We implement this in our PDN via strided average-pooling layers after the first and the second convolutional layer.
With the proposed PDN, we are able to obtain the features for an image of size 256$\times$256 in less than \SI{800}{\micro\second} on an NVIDIA RTX A6000 GPU.

\begin{figure}[t]
\begin{center}
\includegraphics[width=1.0\linewidth]{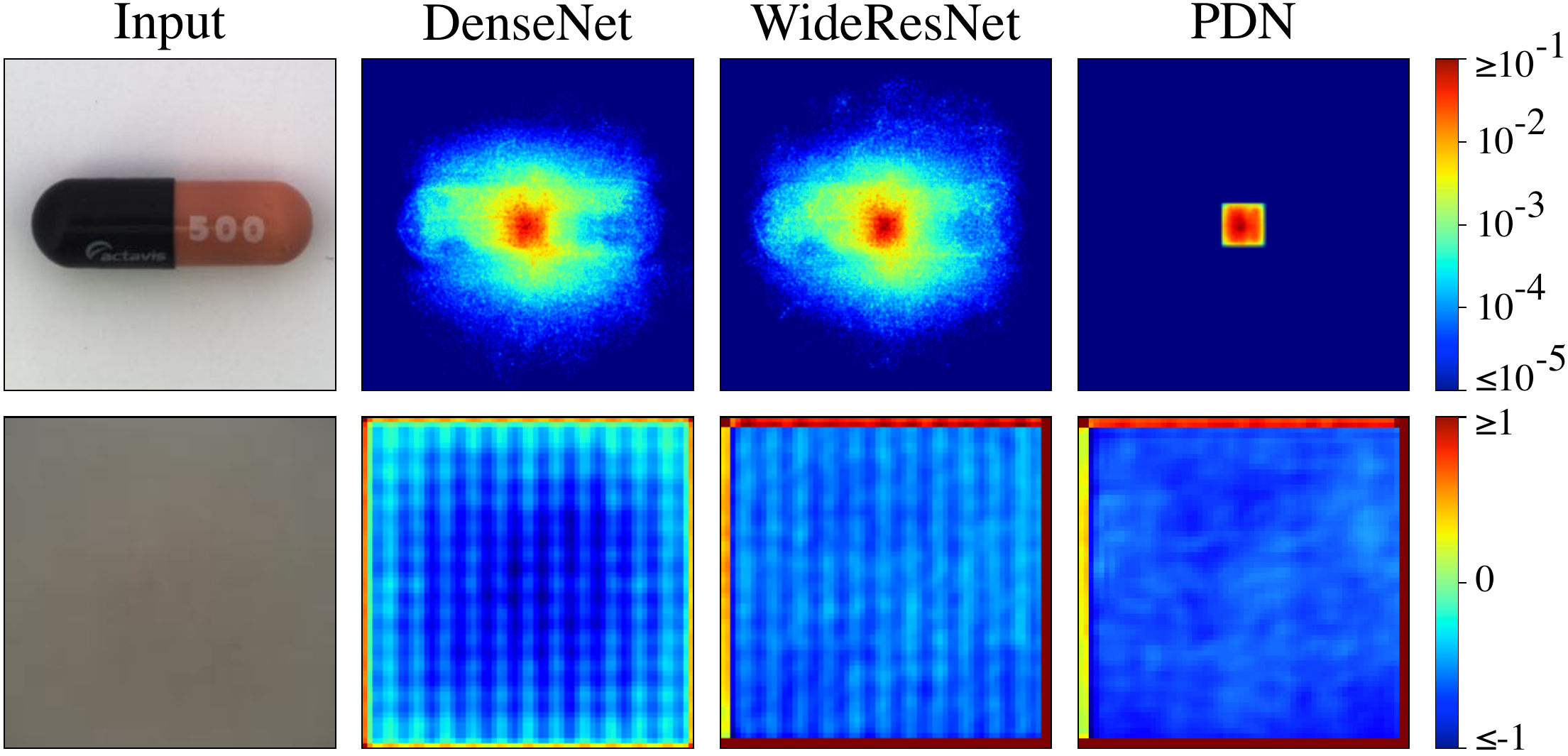}
\end{center}
   \caption{Upper row: absolute gradient of a single feature vector, located in the center of the output, with respect to each input pixel, averaged across input and output channels.
   Lower row: Average feature map of the first output channel across 1000 randomly chosen images from ImageNet \cite{russakovsky2015_alexnet}.
   The mean of these images is shown on the left.
   The feature maps of the DenseNet \cite{huang2017densely} and the WideResNet exhibit strong artifacts.
   }
\label{fig:artifacts}
\vspace*{2mm}
\end{figure}

To make the PDN generate expressive features, we distill a deep pretrained classification network into it.
For a controlled comparison, we use the same pretrained features as PatchCore \cite{roth2022towards} from a WideResNet-101.
We train the PDN on images from ImageNet \cite{russakovsky2015_alexnet} by minimizing the mean squared difference between its output and the features extracted from the pretrained network.
We provide the full list of training hyperparameters in Appendix \ref{subsec:distillation}.
Besides higher efficiency, the PDN has another benefit in comparison to the deep networks used by recent methods.
By design, a feature vector generated by the PDN only depends on the pixels in its respective 33$\times$33 patch.
The feature vectors of pretrained classifiers, on the other hand, exhibit long-range dependencies on other parts of the image.
This is shown in \Cref{fig:artifacts}, using PatchCore's feature extractors as an example.
The well-defined receptive field of the PDN ensures that an anomaly in one part of the image cannot trigger anomalous feature vectors in other, distant parts, which would impair the localization of anomalies.

\subsection{Lightweight Student--Teacher}
\label{sec:method_structural}

For detecting anomalous feature vectors, we use a student--teacher (S--T) approach in which the teacher is given by our distilled PDN.
Since we can execute the PDN in under a millisecond, we use its architecture for the student as well, resulting in a low overall latency.
This lightweight student--teacher pair, however, lacks techniques used by previous methods to increase the anomaly detection performance: ensembling multiple teachers and students \cite{bergmann2020_uninformed_cvpr}, using features from a pyramid of layers \cite{wang2021student_teacher}, and using architectural asymmetry between the student and the teacher network \cite{rudolph2023asymmetric}.
We therefore introduce a training loss that substantially improves the detection of anomalies without affecting the computational requirements at test time.

We observe that in the standard \studteach\ framework, increasing the number of training images can improve the student's ability to imitate the teacher on anomalies.
This worsens the anomaly detection performance.
At the same time, deliberately decreasing the number of training images can suppress important information about normal images.
Our goal is to show the student enough data so that it can mimic the teacher sufficiently on normal images while avoiding generalization to anomalous images.
Similar to Online Hard Example Mining \cite{shrivastava2016training}, we therefore restrict the student's loss to the most relevant parts of an image.
These are the patches where the student currently mimics the teacher the least.
We propose a hard feature loss, which only uses the output elements with the highest loss for backpropagation.

\begin{figure}[t]
\begin{center}
\includegraphics[width=1.0\linewidth]{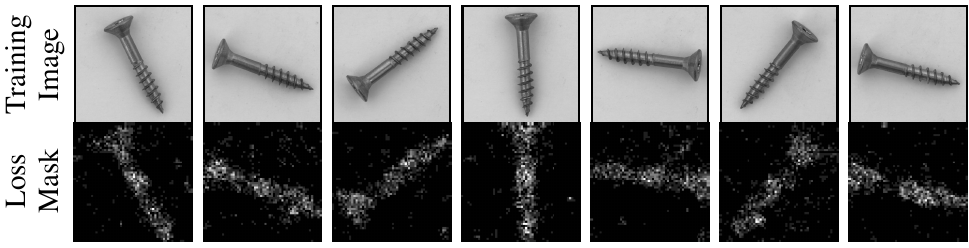}
\end{center}
   \caption{
   Randomly picked loss masks generated by the hard feature loss during training.
   The brightness of a mask pixel indicates how many of the dimensions of the respective feature vector were selected for backpropagation.
   The student network already mimics the teacher well on the background and thus focuses on learning the features of differently rotated screws.
   }
\label{fig:ohem}
\end{figure}

Formally, we apply a teacher $T$ and a student $S$ to a training image $I$, which yields $T(I) \in \mathbb{R}^{C \times W \times H}$ and $S(I) \in \mathbb{R}^{C \times W \times H}$.
We compute the squared difference for each tuple $(c, w, h)$ as $D_{c, w, h} = (T(I)_{c, w, h} - S(I)_{c, w, h})^2$.
Based on a mining factor $p_\mathrm{hard} \in [0, 1]$, we then compute the $p_\mathrm{hard}$-quantile of the elements of $D$.
Given the $p_\mathrm{hard}$-quantile $d_\mathrm{hard}$, we compute the training loss $L_\mathrm{hard}$ as the mean of all $D_{c, w, h} \geq d_\mathrm{hard}$.
Setting $p_\mathrm{hard}$ to zero would yield the original \studteach\ loss.
In our experiments, we set $p_\mathrm{hard}$ to $0.999$, which corresponds to using, on average, ten percent of the values in each of the three dimensions of $D$ for backpropagation.
\Cref{fig:ohem} visualizes the effect of the hard feature loss for $p_\mathrm{hard} = 0.999$.
During inference, the 2D anomaly score map $M \in \mathbb{R}^{W \times H}$ is given by $M_{w, h} = C^{-1} \sum_c D_{c, w, h}$, i.e., by $D$ averaged across channels.
It assigns an anomaly score to each feature vector.
By using the hard feature loss, we avoid outliers in the anomaly scores on normal images, i.e., false-positive detections.

\begin{figure*}[t]
\begin{center}
\includegraphics[width=1.0\linewidth]{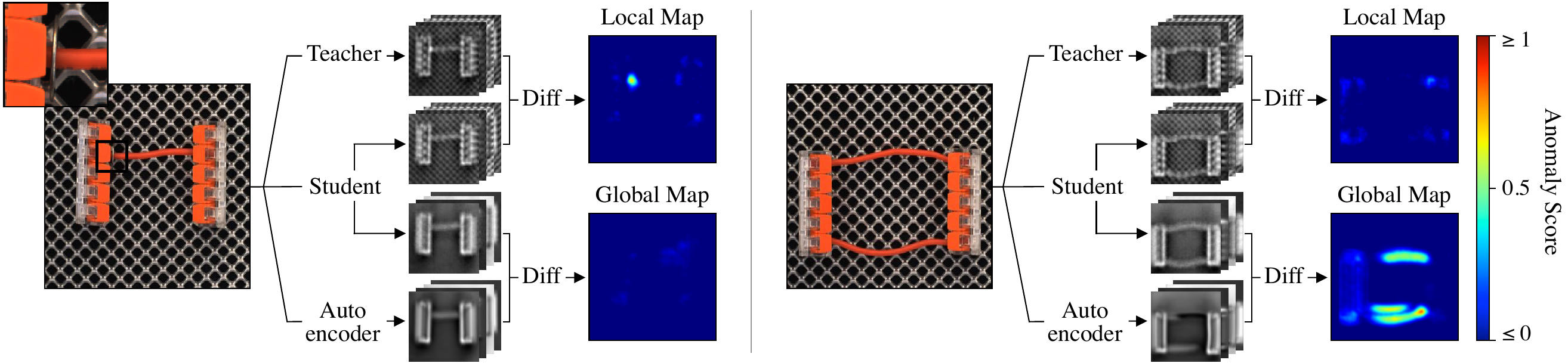}
\end{center}
   \caption{\ourmethod\ applied to two test images from MVTec LOCO\@.
   Normal input images contain a horizontal cable connecting the two splicing connectors at an arbitrary height.
   The anomaly on the left is a foreign object in the form of a small metal washer at the end of the cable.
   It is visible in the local anomaly map because the outputs of the student and the teacher differ.
   The logical anomaly on the right is the presence of a second cable.
   The autoencoder fails to reconstruct the two cables on the right in the feature space of the teacher.
   The student also predicts the output of the autoencoder in addition to that of the teacher.
   Because its receptive field is restricted to small patches of the image, it is not influenced by the presence of the additional red cable.
   This causes the outputs of the autoencoder and the student to differ.
   ``Diff'' refers to computing the element-wise squared difference between two collections of output feature maps and computing its average across feature maps.
   To obtain pixel anomaly scores, the anomaly maps are resized to match the input image using bilinear interpolation.
   }
\label{fig:architecture}
\end{figure*}

In addition to the hard feature loss, we use a loss penalty during training that further hinders the student from imitating the teacher on images that are not part of the normal training images.
In the standard \studteach\ framework, the teacher is pretrained on an image classification dataset, or it is a distilled version of such a pretrained network.
The student is not trained on that pretraining dataset but only on the application's normal images.
We propose to also use the images from the teacher's pretraining during the training of the student.
Specifically, we sample a random image $P$ from the pretraining dataset, in our case ImageNet, in each training step.
We compute the loss of the student as $L_\mathrm{ST} = L_\mathrm{hard} + (CWH)^{-1}\sum_c \|S(P)_c\|_F^2$.
This penalty hinders the student from generalizing its imitation of the teacher to out-of-distribution images.

\subsection{Logical Anomaly Detection}
\label{sec:method_logical}

There are many types of logical anomalies, such as missing, misplaced, or surplus objects or the violation of geometrical constraints, for example, the length of a screw.
As recommended by the authors of the MVTec LOCO dataset \cite{bergmann2021_mvtec_loco_ijcv}, we use an autoencoder for learning logical constraints of the training images and detecting violations of these constraints.
\Cref{fig:architecture} depicts the anomaly detection methodology for \ourmethod.
It consists of the aforementioned student--teacher pair and an autoencoder.
The autoencoder is trained to predict the output of the teacher.
Formally, we apply an autoencoder $A$ to a training image $I$, yielding $A(I) \in \mathbb{R}^{C \times W \times H}$, and compute the loss as $L_\mathrm{AE} = (CWH)^{-1} \sum_c \|T(I)_c - A(I)_c\|_F^2$.
We use a standard convolutional autoencoder comprising strided convolutions in the encoder and bilinear upsampling in the decoder.
We provide the detailed hyperparameters of its layers in in Appendix \ref{sec:details_ours}.

In contrast to the patch-based student, the autoencoder must encode and decode the complete image through a bottleneck of 64 latent dimensions.
On images with logical anomalies, the autoencoder usually fails to generate the correct latent code for reconstructing the image in the teacher's feature space.
However, its reconstructions are also flawed on normal images, as autoencoders generally struggle with reconstructing fine-grained patterns \cite{brox2016_learned_visual_similarity_metrics, bergmann2018_ssim_ae}.
This is the case for the background grids in \Cref{fig:architecture}.
Using the difference between the teacher's output and the autoencoder's reconstruction as an anomaly map would cause false-positive detections in these cases.
Instead, we double the number of output channels of our student network and train it to predict the output of the autoencoder in addition to the output of the teacher.
Let $S'(I) \in \mathbb{R}^{C \times W \times H}$ denote the additional output channels of the student.
The student's additional loss is then $L_\mathrm{STAE} = (CWH)^{-1} \sum_c \|A(I)_c - S'(I)_c\|_F^2$.
The total training loss is the sum of $L_\mathrm{AE}$, $L_\mathrm{ST}$, and $L_\mathrm{STAE}$.

The student learns the systematic reconstruction errors of the autoencoder on normal images, e.g., blurry reconstructions.
At the same time, it does not learn the reconstruction errors for anomalies because these are not part of the training set.
This makes the difference between the autoencoder's output and the student's output well-suited for computing the anomaly map.
Analogous to the student--teacher pair, the anomaly map is the squared difference between the two outputs, averaged across channels.
We refer to this anomaly map as the global anomaly map and to the anomaly map generated by the student--teacher pair as the local anomaly map.
We average these two anomaly maps to compute the combined anomaly map and use its maximum value as the image-level anomaly score.
The combined anomaly map thus contains the detection results of the student--teacher pair and the detection results of the autoencoder--student pair.
Sharing the student's hidden layers in the computation of these detection results allows our method to maintain low computational requirements, while enabling the detection of structural and logical anomalies.

\subsection{Anomaly Map Normalization}
\label{sec:method_balancing}

The local and the global anomaly map must be normalized to similar scales before averaging them to obtain the combined anomaly map.
This is important for cases where the anomaly is only detected in one of the maps, such as in \Cref{fig:architecture}.
Otherwise, noise in one map could make accurate detections in the other map indiscernible in the combined map.
To estimate the scale of the noise in normal images, we use validation images, i.e., unseen images from the training set.
For each of the two anomaly map types, we compute the set of all pixel anomaly scores across the validation images.
We then compute two $p$-quantiles for each set: $q_a$ and $q_b$, for $p=a$ and $p=b$, respectively.
We determine a linear transformation that maps $q_a$ to an anomaly score of $0$ and $q_b$ to a score of $0.1$.
At test time, the local and global anomaly maps are normalized with the respective linear transformation.

By using quantiles, the normalization becomes robust to the distribution of anomaly scores on normal images, which can vary between scenarios.
Whether the scores between~$q_a$ and~$q_b$ are normally distributed or a mixture of Gaussians or follow another distribution has no influence on the normalization.
Our experiments include an ablation study on the values of~$a$ and~$b$.
The choice of the mapping destination values $0$ and $0.1$ has no effect on anomaly detection metrics such as the area under the ROC curve (AU-ROC).
That is because the AU-ROC only depends on the ranking of scores, not on their scale.
We choose $0$ and $0.1$ because they yield maps that are suitable for a standard zero-to-one color scale.

\section{Experiments}

We compare \ourmethod\ to AST \cite{rudolph2023asymmetric}, DSR \cite{zavrtanik2022dsr}, FastFlow \cite{yu2021fastflow}, GCAD \cite{bergmann2021_mvtec_loco_ijcv}, PatchCore \cite{roth2022towards}, SimpleNet \cite{Liu_2023_CVPR}, and \studteach\ \cite{bergmann2020_uninformed_cvpr}, using official implementations where available.
We provide configuration details for all evaluated methods in Appendix \ref{sec:details_others}.
GCAD consists of an ensemble of two anomaly detection models that use different feature extractors.
We find that one of the two ensemble members performs better on average than the combined ensemble and therefore report the results for this member.
This reduces the latency reported for GCAD by a factor of two.
For SimpleNet, we are able to reproduce the official results but find that SimpleNet tunes the training duration on the test images of a scenario.
During training, the model is repeatedly evaluated on all test images and the maximum of all obtained test scores is reported after training.
We disable this technique, since it overestimates the actual performance of the model on unseen images.
In practice, it would furthermore require a validation set with anomalous images.
\mbox{MVTec AD}, VisA, and MVTec LOCO do not include anomalous images in their training and validation sets to avoid defect-type-specific tuning of hyperparameters.
For SimpleNet, we therefore evaluate the final trained model, following common practice.

For PatchCore, we include two variants: the default single model variant, for which the authors report the lowest latency, and the ensemble variant, denoted by \patchcoreens.
We are able to reproduce the official results but disable the cropping of the center \SI{76.6}{\percent} of input images for a fair comparison.
In the case of MVTec AD, \SI{99.9}{\percent} of the defects lie fully or partially within this cropped area.
In real-world applications, anomalies can occur outside of this area as well.
We disable custom cropping, as it implies knowledge about the anomalies in the test set.
For FastFlow, we use the version based on the WideResNet-50-2 feature extractor, as it is similar to the WideResNet used by PatchCore, SimpleNet, and our method.
We use the implementation provided by the Intel anomalib \cite{akcay2022anomalib} but disable early stopping, i.e., the scenario-specific tuning of the training duration on test images, analogously to SimpleNet.
With early stopping enabled, EfficientAD itself achieves an image-level detection AU-ROC of \SI{99.8}{\percent} on MVTec AD.

For our method, we evaluate two variants: \mbox{\ourmethod-S} and \ourmethod-M.
\ourmethod-S uses the architecture displayed in \Cref{fig:pdn} for the teacher and the student.
For \ourmethod-M, we double the number of kernels in the hidden convolutional layers of the teacher and the student.
Furthermore, we insert a 1$\times$1 convolution after the second pooling layer and after the last convolutional layer.
We provide a list of implementation details, such as the learning rate schedule, in Appendix \ref{subsec:training_and_evaluation}.

We evaluate each method on the 32 anomaly detection scenarios of MVTec AD, VisA, and MVTec LOCO\@.
The anomaly detection performance of a method is measured with the AU-ROC based on its predicted image-level anomaly scores.
We measure the anomaly localization performance using the AU-PRO segmentation metric up to a false positive rate of \SI{30}{\percent}, as recommended by \cite{bergmann2021_mvtec_ad_ijcv}.
For MVTec LOCO, we use the AU-sPRO metric \cite{bergmann2021_mvtec_loco_ijcv}, a generalization of the AU-PRO metric for evaluating the localization of logical anomalies.
Appendix \ref{sec:ad_metrics} provides the results for additional anomaly detection metrics, such as the area under the precision-recall curve and the pixel-wise AU-ROC computed on pixel anomaly scores.

\addtolength{\tabcolsep}{-2pt}  
\begin{table}
\small
\begin{center}
\begin{tabular}{ccccc}
Method & \specialcell[c]{Detect. \\ AU-ROC} & \specialcell[c]{Segment. \\ AU-PRO} & \specialcell[c]{Latency \\ {[ms]}} & \specialcell[c]{Throughput \\ {[img / s]}} \\
\hline
GCAD & 85.4 & 88.0 & 11 & 121 \Tstrut\\
SimpleNet & 87.9 & 74.4 & 12 & 194 \\
\studteach & 88.4 & 89.7 & 75 & 16\\
FastFlow & 90.0 & 86.5 & 17 & 120\\
DSR & 90.8 & 78.6 & 17 & 104 \\
PatchCore & 91.1 & 80.9 & 32 & 76 \\
\patchcoreens & 92.1 & 80.7 & 148 & 13 \\
AST & 92.4 & 77.2 & 53 & 41  \Bstrut\\
\hline
\ourmethod-S & \meanwithstd{95.4}{0.06} & \meanwithstd{92.5}{0.05} & \meanwithstd{\textbf{2.2}}{0.01} & \meanwithstd{\textbf{614}}{2} \rule{0pt}{3.6ex}\\
\ourmethod-M & \meanwithstd{\textbf{96.0}}{0.09} & \meanwithstd{\textbf{93.3}}{0.04} & \meanwithstd{4.5}{0.01} & \meanwithstd{269}{1} \\
\end{tabular}
\end{center}
\caption{Anomaly detection and anomaly localization performance in comparison to the latency and throughput. Each AU-ROC and AU-PRO percentage is an average of the mean AU-ROCs and mean AU-PROs, respectively, on MVTec AD, VisA, and MVTec LOCO.
For \ourmethod, we report the mean and standard deviation of five runs.
}
\label{tab:main}
\end{table}
\addtolength{\tabcolsep}{2pt}

\addtolength{\tabcolsep}{-3pt}  
\begin{table}
\small
\begin{center}
\begin{tabular}{ccccc||cc}
Method & MAD & LOCO & VisA & Mean & \specialcell[c]{LOCO \\ Logic.} & \specialcell[c]{LOCO \\ Struct.} \\
\hline
GCAD & 89.1 & 83.3 & 83.7 & 85.4 & 83.9 & 82.7 \Tstrut\\
SimpleNet & 98.2 & 77.6 & 87.9 & 87.9 & 71.5 & 83.7 \\
\studteach & 93.2 & 77.4 & 94.6 & 88.4 & 66.5 & 88.3 \\
FastFlow & 96.9 & 79.2 & 93.9 & 90.0 & 75.5 & 82.9 \\
DSR & 98.1 & 82.6 & 91.8 & 90.8 & 75.0 & 90.2 \\
PatchCore & 98.7 & 80.3 & 94.3 & 91.1 & 75.8 & 84.8 \\
\patchcoreens & \textbf{99.3} & 79.4 & 97.7 & 92.1 & 71.0 & 87.7 \\
AST & 98.9 & 83.4 & 94.9 & 92.4 & 79.7 & 87.1 \Bstrut\\
\hline
\ourmethod-S & 98.8 & 90.0 & 97.5 & 95.4 & 85.8 & 94.1 \Tstrut\\
\ourmethod-M & 99.1 & \textbf{90.7} & \textbf{98.1} & \textbf{96.0} & \textbf{86.8} & \textbf{94.7} \\
\end{tabular}
\end{center}
\caption{Mean anomaly detection AU-ROC percentages per dataset collection (left) and on the logical and structural anomalies of MVTec LOCO (right). For \ourmethod, we report the mean of five runs.
Performing method development solely on MVTec AD (MAD) becomes prone to overfitting design choices to the few remaining misclassified test images.}
\label{tab:per_dataset}
\end{table}
\addtolength{\tabcolsep}{3pt}

\addtolength{\tabcolsep}{-0.5pt}  
\begin{table}
\small
\begin{center}
\begin{tabular}{ccccccc}
$a$ (for $q_a$) & 0.5 & 0.8 & \textbf{0.9} & 0.95 & 0.98 & 0.99 \Bstrut\\
\hline
AU-ROC & 95.9 & 95.9 & 96.0 & 95.9 & 95.9 & 95.8 \Tstrut\\
\rule{0pt}{5mm}
$b$ (for $q_b$) & 0.95 & 0.98 & 0.99 & \textbf{0.995} & 0.998 & 0.999 \Bstrut\\
\hline
AU-ROC & 95.8 & 95.9 & 96.0 & 96.0 & 95.9 & 95.9 \Tstrut\\
\rule{0pt}{5mm} 
$p_\mathrm{hard}$ & 0 & 0.9 & 0.99 & \textbf{0.999} & 0.9999 & 0.99999 \Bstrut\\
\hline
AU-ROC & 94.9 & 94.9 & 95.7 & 96.0 & 95.8 & 95.7 \Tstrut\\
\end{tabular}
\end{center}
\caption{Mean anomaly detection AU-ROC of \mbox{\ourmethod-M} on MVTec AD, VisA, and MVTec LOCO when varying the locations of quantiles.
These are the two sampling points $a$ and $b$ of the quantile-based map normalization and the mining factor $p_\mathrm{hard}$.
Setting $p_\mathrm{hard}$ to zero disables the proposed hard feature loss.
Default values used in our experiments are highlighted in bold.
}
\label{tab:hyperparams}
\vspace*{-8mm}
\end{table}
\addtolength{\tabcolsep}{-0.5pt}

When reporting the AU-ROC or AU-PRO for a dataset collection, we follow the policy of the dataset authors.
For each collection, we evaluate the respective metric for each scenario and then compute the mean across scenarios.
For MVTec LOCO, we use the official evaluation script, which gives logical and structural anomalies an equal weight in the computed metrics.
When reporting the average AU-ROC or AU-PRO on the three dataset collections, we compute the average of the three dataset means.
Thus, an overall average score weights logical anomalies and structural anomalies by roughly one-sixth and five-sixths, respectively.
We provide the evaluation results for each of the 32 anomaly detection scenarios individually in the appendix to enable an evaluation with a custom weighting.

\Cref{tab:main} reports the overall anomaly detection performance for each method.
\ourmethod\ achieves a strong image-level detection and pixel-level localization of anomalies.
Reliably localizing anomalies in an image provides explainable detection results and allows the discovery of spurious correlations in detections.
It also enables a flexible postprocessing, such as excluding defect segmentations based on their size.

\Cref{tab:per_dataset} breaks down the overall anomaly detection performance into the three dataset collections.
It shows that the lead of \ourmethod\ on MVTec LOCO is in equal parts due to its performance on logical and on structural anomalies.
In \Cref{tab:hyperparams}, we assess the robustness of \ourmethod\ to varying hyperparameters.

\begin{figure}[t]
\begin{center}
\includegraphics[width=1.0\linewidth]{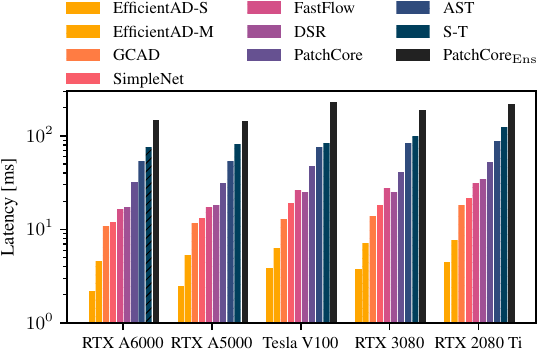}
\end{center}
\vspace*{3mm}
   \caption{
   Latency per GPU. The ranking of methods is the same on each GPU, except for two cases in which DSR is slightly faster than FastFlow.}
\label{fig:latency_per_gpu}
\vspace*{4mm}
\end{figure}

Furthermore, we measure the computational cost of each method during inference.
As explained above, the number of parameters can be a misleading proxy metric for the latency and throughput of convolutional architectures since it does not consider the resolution of a convolution's input feature map, i.e., how often a parameter is used in a forward pass.
Similarly, the number of floating point operations (FLOPs) can be misleading since it does not take into account how easily computations can be parallelized.
For transparency, we report the number of parameters, the number of FLOPs, and the memory footprint of each method in Appendix \ref{sec:efficiency_metrics}.
Here, we focus on the metrics that are most relevant in anomaly detection applications: the latency and the throughput.
We measure the latency with a batch size of 1 and the throughput with a batch size of 16.
\Cref{tab:main} reports the measurements for each method on an NVIDIA RTX A6000 GPU.
\Cref{fig:latency_per_gpu} shows the latency of each method on each of the GPUs in our experimental setup.
Appendix \ref{sec:efficiency_metrics} contains a detailed description of our timing methodology.

\begin{figure}[t]
\begin{center}
\includegraphics[width=1.0\linewidth]{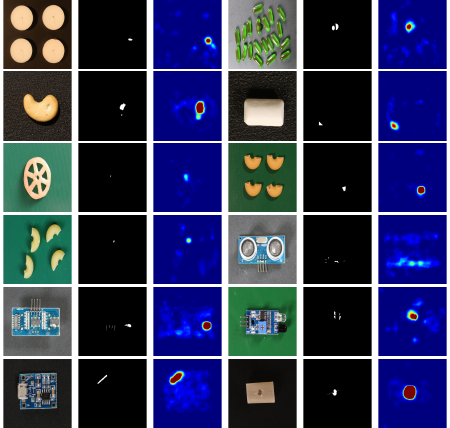}
\end{center}
\vspace*{-1mm}
   \caption{Non-cherry-picked qualitative results of \ourmethod\ on VisA. For each of its 12 scenarios, we show a randomly sampled defect image, the ground truth segmentation mask, and the anomaly map generated by EfficientAD-M.}
\label{fig:qual}
\end{figure}

In \Cref{fig:qual}, we show randomly sampled qualitative results of \ourmethod\ on the VisA dataset collection.
Appendix \ref{sec:qualitative_results} provides qualitative results for the other evaluated methods and dataset collections as well.

We examine the effects of the components of \ourmethod\ in the ablation study shown in \Cref{tab:ablation_cumulative} and \Cref{tab:ablation_isolated}.
For experiments without the proposed quantile-based map normalization, we use a Gaussian-based map normalization as a baseline instead.
There, we compute the linear transformation parameters such that pixel anomaly scores on the validation set have a mean of zero and a variance of one.
This baseline normalization is sensitive to the distribution of validation anomaly scores, which can vary between scenarios.
The quantile-based normalization is independent of how the scores between $q_a$ and $q_b$ are distributed and performs substantially better than the baseline.

We also evaluate the effect of the two proposed loss terms for training the student--teacher pair.
The hard feature loss increases the anomaly detection AU-ROC by \SI{1.0}{\percent} in \Cref{tab:ablation_cumulative}.
This improvement alone is greater than or equal to each of the improvement margins between the consecutive rows of FastFlow, DSR, PatchCore, \patchcoreens, and AST in \Cref{tab:main}.
The student's penalty on pretraining images further improves the anomaly detection performance.
Notably, the proposed map normalization, the hard feature loss, and the pretraining penalty keep the computational requirements of \ourmethod\ low, while creating a substantial margin w.r.t. the anomaly detection performance.

\begin{table}
\small
\begin{center}
\begin{tabular}{lccc}
 & \specialcell[c]{Detection \\ AU-ROC} & Diff. & \specialcell[c]{Latency \\ {[ms]}} \\
\hline
PDN & 93.2 & & 2.2 \Tstrut\\
$\hookrightarrow$ with map normalization & 94.0 & + 0.8 & 2.2 \\
\, $\hookrightarrow$ with hard feature loss & 95.0 & + 1.0  & 2.2 \\
\, \, $\hookrightarrow$ with pretraining penalty & 95.4 & + 0.4 & 2.2 \Bstrut\\
\hline
\ourmethod-S & 95.4 & & 2.2 \\
\ourmethod-M & 96.0 & + 0.6 & 4.5 \\
\end{tabular}
\end{center}
\caption{Cumulative ablation study in which techniques are gradually combined to form EfficientAD. Each AU-ROC percentage is an average of the mean AU-ROCs on MVTec AD, VisA, and MVTec LOCO.}
\label{tab:ablation_cumulative}
\end{table}

\begin{table}
\small
\begin{center}
\begin{tabular}{lccc}
 & \specialcell[c]{Detection \\ AU-ROC} & Diff. & \specialcell[c]{Latency \\ {[ms]}} \\
\hline
\ourmethod-S & 95.4 & & 2.2 \Tstrut\\
Without map normalization \, \, & 94.7 & - 0.7 & 2.2 \\
Without hard feature loss & 94.7 & - 0.7 & 2.2 \\
Without pretraining penalty & 95.0 & - 0.4 & 2.2 \\
\end{tabular}
\end{center}
\caption{Isolated ablation study in which techniques are separately removed from EfficientAD-S.}
\label{tab:ablation_isolated}
\vspace*{1mm}
\end{table}

\section{Conclusion}

In this paper, we introduce \ourmethod, a method with a strong anomaly detection performance and a high computational efficiency.
It sets new standards for the detection of structural as well as logical anomalies.
Both \ourmethod-S and \ourmethod-M outperform other methods on the detection and the localization of anomalies by a large margin.
Compared to AST, the second-best method, \ourmethod-S reduces the latency by a factor of 24 and increases the throughput by a factor of 15.
Its low latency, high throughput, and high detection rate make it suitable for real-world applications.
For future anomaly detection research, \ourmethod\ is an important baseline and a fruitful foundation.
Its efficient patch description network, for instance, can be used as a feature extractor in other anomaly detection methods as well to reduce their latency.

\paragraph{Limitations.}
The student--teacher model and the autoencoder are designed to detect anomalies of different types.
The autoencoder detects logical anomalies, while the student--teacher model detects coarse and fine-grained structural anomalies.
Fine-grained logical anomalies, however, remain a challenge -- for example a screw that is two millimeters too long.
To detect these, practitioners would have to use traditional metrology methods \cite{steger2018_mva_book}.
As for the limitations in comparison to other recent anomaly detection methods:
In contrast to kNN-based methods, our approach requires training, especially for the autoencoder to learn the logical constraints of normal images.
This takes twenty minutes in our experimental setup.

\onecolumn

\vspace{-1mm}
\section*{Appendices}

\appendices

We provide the following supplementary material:

\begin{itemize}[topsep=0.1em]
    \itemsep-1mm
    \item (\ref{sec:details_ours}) Implementation details for \ourmethod, including the training and inference procedure of \ourmethod\ on the dataset of an anomaly detection scenario (\ref{subsec:training_and_evaluation}) and the distillation training of the patch description network (\ref{subsec:distillation}).
    \item (\ref{sec:details_others}) Implementation and configuration details for other evaluated methods.
    \item (\ref{sec:backbones}) Evaluation of the anomaly detection performance of \ourmethod\ for different distillation backbones.
    \item (\ref{sec:ad_metrics}) Results for additional anomaly detection metrics, such as the area under the precision recall curve.
    \item (\ref{sec:efficiency_metrics}) Description of our timing methodology and results for additional computational efficiency metrics such as the number of parameters.
    \item (\ref{sec:qualitative_results}) Qualitative results in the form of anomaly maps generated by \ourmethod\ and other methods on the evaluated datasets.
    \item Anomaly detection results for each method on each of the 32 scenarios from MVTec AD, VisA, and MVTec LOCO in the \href{https://www.mydrive.ch/shares/79401/c41dcdb937972fb43d5cdd7bfa7072f8/download/449203483-1690527455/per_scenario_results.json}{\texttt{per\_scenario\_results.json}} file \footnote{\url{https://www.mydrive.ch/shares/79401/c41dcdb937972fb43d5cdd7bfa7072f8/download/449203483-1690527455/per_scenario_results.json}}.
\end{itemize}
\vspace{0cm}

\section{Implementation Details for \ourmethod}
\label{sec:details_ours}

\subsection{Training and Inference}
\label{subsec:training_and_evaluation}

\Cref{alg:training} describes the training of \ourmethod-S and \Cref{alg:inference} explains the inference procedure.
For \mbox{\ourmethod-M}, replace the architecture of \Cref{tab:arch_eads} with that of \Cref{tab:arch_eadm}.

\begin{algorithm}[h!]
  \caption{\ourmethod-S Training Algorithm} 
  \label{alg:training} 
  \begin{algorithmic} [1]
    \Require{A pretrained teacher network $T : \mathbb{R}^{3 \times 256 \times 256} \rightarrow \mathbb{R}^{384 \times 64 \times 64}$ with an architecture as given in \Cref{tab:arch_eads}}
    \Require{A sequence of training images $\mathcal{I}_\mathrm{train}$ with $I_\mathrm{train} \in \mathbb{R}^{3 \times 256 \times 256}$ for each $I_\mathrm{train} \in \mathcal{I}_\mathrm{train}$}
    \Require{A sequence of validation images $\mathcal{I}_\mathrm{val}$ with $I_\mathrm{val} \in \mathbb{R}^{3 \times 256 \times 256}$ for each $I_\mathrm{val} \in \mathcal{I}_\mathrm{val}$}
    \State{Randomly initialize a student network $S : \mathbb{R}^{3 \times 256 \times 256} \rightarrow \mathbb{R}^{768 \times 64 \times 64}$ with an architecture as given in \Cref{tab:arch_eads}}
    \State{Randomly initialize an autoencoder $A : \mathbb{R}^{3 \times 256 \times 256} \rightarrow \mathbb{R}^{384 \times 64 \times 64}$ with an architecture as given in \Cref{tab:arch_ae}}
    \For{$c \in {1, \dots, 384}$} \Comment{Compute teacher channel normalization parameters $\mu \in \mathbb{R}^{384}$ and $\sigma \in \mathbb{R}^{384}$}
    \State{Initialize an empty sequence $X \leftarrow (~)$}
    \For{$I_\mathrm{train} \in \mathcal{I}_\mathrm{train}$}
        \State{$Y'\leftarrow T(I_\mathrm{train})$}
        \State{$X \leftarrow X^\frown \mathrm{vec}(Y'_c)$} \Comment{Append the channel output to $X$}
    \EndFor
    \State{Set $\mu_c$ to the mean and $\sigma_c$ to the standard deviation of the elements of $X$}
    \algstore{alg_training}
	\end{algorithmic}
\end{algorithm}

\begin{algorithm}[h]
  \begin{algorithmic} [1]
    \algrestore{alg_training}
    \EndFor
    \State{Initialize Adam \cite{kingma2014_adam} with a learning rate of $10^{-4}$ and a weight decay of $10^{-5}$ for the parameters of $S$ and $A$}
    \For{iteration $= 1,\dots, \num{70000}$}
        \State{Choose a random training image $I_\mathrm{train}$ from $\mathcal{I}_\mathrm{train}$}
        \State{$Y'\leftarrow T(I_\mathrm{train})$} \Comment{Forward pass of the student--teacher pair}
        \State{Compute the normalized teacher output $\hat{Y}$ given by $\hat{Y}_c = (Y'_c - \mu_c) \sigma_c^{-1}$ for each $c \in \{1, \dots, 384\}$}
        \State{$Y^\mathrm{S} \leftarrow S(I_\mathrm{train})$}
        \State{Set $Y^\mathrm{ST} \in \mathbb{R}^{384 \times 64 \times 64}$ to the first 384 channels of $Y^\mathrm{S} \in \mathbb{R}^{768 \times 64 \times 64}$}
        \State{Compute the squared difference between $\hat{Y}$ and $Y^\mathrm{ST}$ for each tuple $(c, w, h)$ as $D^\mathrm{ST}_{c, w, h} = (\hat{Y}_{c, w, h} - Y^\mathrm{ST}_{c, w, h})^2$}
        \State{Compute the $0.999$-quantile of the elements of $D^\mathrm{ST}$, denoted by $d_\mathrm{hard}$}
        \State{Compute the loss $L_\mathrm{hard}$ as the mean of all $D^\mathrm{ST}_{c, w, h} \geq d_\mathrm{hard}$ }
        \State{Choose a random pretraining image $P \in \mathbb{R}^{3 \times 256 \times 256}$ from ImageNet \cite{russakovsky2015_alexnet}}
        \State{Compute the loss $L_\mathrm{ST} = L_\mathrm{hard} + (384 \cdot 64 \cdot 64)^{-1}\sum_{c=1}^{384} \|S(P)_c\|_F^2$}
        \State{Randomly choose an augmentation index $i_\mathrm{aug} \in \{1, 2, 3\}$} \Comment{Augment $I_\mathrm{train}$ for $A$ using torchvision \cite{paszke2019_PyTorch}}
        \State{Sample an augmentation coefficient $\lambda$ from the uniform distribution $U(0.8, 1.2)$}
        \If{$i_\mathrm{aug} == 1$}
        $I_\mathrm{aug} \leftarrow \mathtt{torchvision.transforms.functional\_pil.adjust\_brightness}(I_\mathrm{train}, \lambda)$
        \ElsIf{$i_\mathrm{aug} == 2$}
            $I_\mathrm{aug} \leftarrow \mathtt{torchvision.transforms.functional\_pil.adjust\_contrast}(I_\mathrm{train}, \lambda)$
        \ElsIf{$i_\mathrm{aug} == 3$}
            $I_\mathrm{aug} \leftarrow \mathtt{torchvision.transforms.functional\_pil.adjust\_saturation}(I_\mathrm{train}, \lambda)$
        \EndIf
        \State{$Y^\mathrm{A} \leftarrow A(I_\mathrm{aug})$} \Comment{Forward pass of the autoencoder--student pair}
        \State{$Y'\leftarrow T(I_\mathrm{aug})$}
        \State{Compute the normalized teacher output $\hat{Y}$ given by $\hat{Y}_c = \sigma_c^{-1}(Y'_c - \mu_c)$ for each $c \in \{1, \dots, 384\}$}
        \State{$Y^\mathrm{S} \leftarrow S(I_\mathrm{aug})$}
        \State{Set $Y^\mathrm{STAE} \in \mathbb{R}^{384 \times 64 \times 64}$ to the last 384 channels of $Y^\mathrm{S} \in \mathbb{R}^{768 \times 64 \times 64}$}
        \State{Compute the squared difference between $\hat{Y}$ and $Y^\mathrm{A}$ for each tuple $(c, w, h)$ as $D^\mathrm{AE}_{c, w, h} = (\hat{Y}_{c, w, h} - Y^\mathrm{A}_{c, w, h})^2$}
        \State{Compute the squared difference between $Y^\mathrm{A}$ and $Y^\mathrm{STAE}$ for each tuple $(c, w, h)$ as $D^\mathrm{STAE}_{c, w, h} = (Y^\mathrm{A}_{c, w, h} - Y^\mathrm{STAE}_{c, w, h})^2$}
        \State{Compute the loss $L_\mathrm{AE}$ as the mean of all elements $D^\mathrm{AE}_{c, w, h}$ of $D^\mathrm{AE}$}
        \State{Compute the loss $L_\mathrm{STAE}$ as the mean of all elements $D^\mathrm{STAE}_{c, w, h}$ of $D^\mathrm{STAE}$}
        \State{Compute the total loss $L_\mathrm{total} = L_\mathrm{ST} + L_\mathrm{AE} + L_\mathrm{STAE}$}  \Comment{Backward pass}
        \State{Update the union of the parameters of $S$ and $A$, denoted by $\phi$, using the gradient $\nabla_\phi L_\mathrm{total}$ }
        \If{iteration $>$ \num{66500}}
        \State{Decay the learning rate to $10^{-5}$}
        \EndIf
    \EndFor
    \State{Initialize empty sequences $X_\mathrm{ST} \leftarrow (~)$ and $X_\mathrm{AE} \leftarrow (~)$} \Comment{Quantile-based map normalization on validation images}
    \For{$I_\mathrm{val} \in \mathcal{I}_\mathrm{val}$}
        \State{$Y'\leftarrow T(I_\mathrm{val}),\;\;Y^\mathrm{S} \leftarrow S(I_\mathrm{val}),\;\;Y^\mathrm{A} \leftarrow A(I_\mathrm{val})$}
        \State{Compute the normalized teacher output $\hat{Y}$ given by $\hat{Y}_c = (Y'_c - \mu_c) \sigma_c^{-1}$ for each $c \in \{1, \dots, 384\}$}
        \State{Split the student output into $Y^\mathrm{ST} \in \mathbb{R}^{384 \times 64 \times 64}$ and $Y^\mathrm{STAE} \in \mathbb{R}^{384 \times 64 \times 64}$ as above}
        \State{Compute the squared difference $D^\mathrm{ST}_{c, w, h} = (\hat{Y}_{c, w, h} - Y^\mathrm{ST}_{c, w, h})^2$ for each tuple $(c, w, h)$}
        \State{Compute the squared difference $D^\mathrm{STAE}_{c, w, h} = (Y^\mathrm{A}_{c, w, h} - Y^\mathrm{STAE}_{c, w, h})^2$ for each tuple $(c, w, h)$ }
        \State{Compute the anomaly maps $M_\mathrm{ST} = 384^{-1}\sum_{c=1}^{384} D^\mathrm{ST}_c$ and $M_\mathrm{AE} = 384^{-1}\sum_{c=1}^{384} D^\mathrm{STAE}_c$}
        \State{Resize $M_\mathrm{ST}$ and $M_\mathrm{AE}$ to $256 \times 256$ pixels using bilinear interpolation}
        \State{$X_\mathrm{ST} \leftarrow {X_\mathrm{ST}}^\frown \mathrm{vec}(M_\mathrm{ST})$} \Comment{Append to the sequence of local anomaly scores}
        \State{$X_\mathrm{AE} \leftarrow {X_\mathrm{AE}}^\frown \mathrm{vec}(M_\mathrm{AE})$} \Comment{Append to the sequence of global anomaly scores}
    \EndFor
    \State{Compute the $0.9$-quantile $q_a^\mathrm{ST}$ and the $0.995$-quantile $q_b^\mathrm{ST}$ of the elements of $X_\mathrm{ST}$.}
    \State{Compute the $0.9$-quantile $q_a^\mathrm{AE}$ and the $0.995$-quantile $q_b^\mathrm{AE}$ of the elements of $X_\mathrm{AE}$.}
    \State \Return{$T$, $S$, $A$, $\mu$, $\sigma$, $q_a^\mathrm{ST}$, $q_b^\mathrm{ST}$, $q_a^\mathrm{AE}$, and $q_b^\mathrm{AE}$ }
	\end{algorithmic}
\end{algorithm}

\begin{algorithm}
  \caption{\ourmethod\  Inference Procedure} 
  \label{alg:inference} 
  \begin{algorithmic} [1]
    \Require{$T$, $S$, $A$, $\mu$, $\sigma$, $q_a^\mathrm{ST}$, $q_b^\mathrm{ST}$, $q_a^\mathrm{AE}$, and $q_b^\mathrm{AE}$, as returned by \Cref{alg:training}}
    \Require{Test image $I_\mathrm{test} \in \mathbb{R}^{3 \times 256 \times 256}$}
    \State{$Y'\leftarrow T(I_\mathrm{test}),\;\;Y^\mathrm{S} \leftarrow S(I_\mathrm{test}),\;\;Y^\mathrm{A} \leftarrow A(I_\mathrm{test})$}
    \State{Compute the normalized teacher output $\hat{Y}$ given by $\hat{Y}_c = (Y'_c - \mu_c) \sigma_c^{-1}$ for each $c \in \{1, \dots, 384\}$}
    \State{Split the student output into $Y^\mathrm{ST} \in \mathbb{R}^{384 \times 64 \times 64}$ and $Y^\mathrm{STAE} \in \mathbb{R}^{384 \times 64 \times 64}$ as above}
    \State{Compute the squared difference $D^\mathrm{ST}_{c, w, h} = (\hat{Y}_{c, w, h} - Y^\mathrm{ST}_{c, w, h})^2$ for each tuple $(c, w, h)$}
    \State{Compute the squared difference $D^\mathrm{STAE}_{c, w, h} = (Y^\mathrm{A}_{c, w, h} - Y^\mathrm{STAE}_{c, w, h})^2$ for each tuple $(c, w, h)$ }
    \State{Compute the anomaly maps $M_\mathrm{ST} = 384^{-1}\sum_{c=1}^{384} D^\mathrm{ST}_c$ and $M_\mathrm{AE} = 384^{-1}\sum_{c=1}^{384} D^\mathrm{STAE}_c$}
    \State{Resize $M_\mathrm{ST}$ and $M_\mathrm{AE}$ to $256 \times 256$ pixels using bilinear interpolation}
    \State{Compute the normalized $\hat{M}_\mathrm{ST} = 0.1 (M_\mathrm{ST} - q_a^\mathrm{ST}) (q_b^\mathrm{ST} - q_a^\mathrm{ST})^{-1}$}
    \State{Compute the normalized $\hat{M}_\mathrm{AE} = 0.1 (M_\mathrm{AE} - q_a^\mathrm{AE}) (q_b^\mathrm{AE} - q_a^\mathrm{AE})^{-1}$}
    \State{Compute the combined anomaly map $M = 0.5 \hat{M}_\mathrm{ST} + 0.5 \hat{M}_\mathrm{AE}$}
    \State{Compute the image-level score as $m_\mathrm{image} = \max_{i, j} M_{i, j}$}
    \State \Return{$M$ and $m_\mathrm{image}$}
	\end{algorithmic}
\end{algorithm}

\begin{table}
\begin{center}
\begin{tabular}{cccccc}
Layer Name & Stride & Kernel Size & Number of Kernels & Padding & Activation \\
\hline
Conv-1 & 1$\times$1 & 4$\times$4 & 128 & 3 & ReLU \\
AvgPool-1 & 2$\times$2 & 2$\times$2 & 128 & 1 & - \\
Conv-2 & 1$\times$1 & 4$\times$4 & 256 & 3 & ReLU \\
AvgPool-2 & 2$\times$2 & 2$\times$2 & 256 & 1 & - \\
Conv-3 & 1$\times$1 & 3$\times$3 & 256 & 1 & ReLU \\
Conv-4 & 1$\times$1 & 4$\times$4 & 384 & 0 & - \\
\hline
\end{tabular}
\end{center}
\caption{Patch description network architecture of the teacher network for \ourmethod-S. The student network has the same architecture, but 768 kernels instead of 384 in the Conv-4 layer. A padding value of 3 means that three rows, or columns respectively, of zeros are appended at each border of an input feature map.}
\label{tab:arch_eads}
\end{table}

\begin{table}
\begin{center}
\begin{tabular}{cccccc}
Layer Name & Stride & Kernel Size & Number of Kernels & Padding & Activation \\
\hline
Conv-1 & 1$\times$1 & 4$\times$4 & 256 & 3 & ReLU \\
AvgPool-1 & 2$\times$2 & 2$\times$2 & 256 & 1 & - \\
Conv-2 & 1$\times$1 & 4$\times$4 & 512 & 3 & ReLU \\
AvgPool-2 & 2$\times$2 & 2$\times$2 & 512 & 1 & - \\
Conv-3 & 1$\times$1 & 1$\times$1 & 512 & 0 & ReLU \\
Conv-4 & 1$\times$1 & 3$\times$3 & 512 & 1 & ReLU \\
Conv-5 & 1$\times$1 & 4$\times$4 & 384 & 0 & ReLU \\
Conv-6 & 1$\times$1 & 1$\times$1 & 384 & 0 & - \\
\hline
\end{tabular}
\end{center}
\caption{Patch description network architecture of the teacher network for \ourmethod-M. The student network has the same architecture, but 768 kernels instead of 384 in the Conv-5 and Conv-6 layers. A padding value of 3 means that three rows, or columns respectively, of zeros are appended at each border of an input feature map.}
\label{tab:arch_eadm}
\end{table}

\begin{table}
\begin{center}
\begin{tabular}{cccccc}
Layer Name & Stride & Kernel Size & Number of Kernels & Padding & Activation \\
\hline
EncConv-1 & 2$\times$2 & 4$\times$4 & 32 & 1 & ReLU \\
EncConv-2 & 2$\times$2 & 4$\times$4 & 32 & 1 & ReLU \\
EncConv-3 & 2$\times$2 & 4$\times$4 & 64 & 1 & ReLU \\
EncConv-4 & 2$\times$2 & 4$\times$4 & 64 & 1 & ReLU \\
EncConv-5 & 2$\times$2 & 4$\times$4 & 64 & 1 & ReLU \\
EncConv-6 & 1$\times$1 & 8$\times$8 & 64 & 0 & - \\
Bilinear-1 & \multicolumn{5}{c}{Resizes the 1$\times$1 input features maps to 3$\times$3}  \\
DecConv-1 & 1$\times$1 & 4$\times$4 & 64 & 2 & ReLU \\
Dropout-1 & \multicolumn{5}{c}{Dropout rate = 0.2} \\
Bilinear-2 & \multicolumn{5}{c}{Resizes the 4$\times$4 input features maps to 8$\times$8}  \\
DecConv-2 & 1$\times$1 & 4$\times$4 & 64 & 2 & ReLU \\
Dropout-2 & \multicolumn{5}{c}{Dropout rate = 0.2} \\
Bilinear-3 & \multicolumn{5}{c}{Resizes the 9$\times$9 input features maps to 15$\times$15}  \\
DecConv-3 & 1$\times$1 & 4$\times$4 & 64 & 2 & ReLU \\
Dropout-3 & \multicolumn{5}{c}{Dropout rate = 0.2} \\
Bilinear-4 & \multicolumn{5}{c}{Resizes the 16$\times$16 input features maps to 32$\times$32}  \\
DecConv-4 & 1$\times$1 & 4$\times$4 & 64 & 2 & ReLU \\
Dropout-4 & \multicolumn{5}{c}{Dropout rate = 0.2} \\
Bilinear-5 & \multicolumn{5}{c}{Resizes the 33$\times$33 input features maps to 63$\times$63}  \\
DecConv-5 & 1$\times$1 & 4$\times$4 & 64 & 2 & ReLU \\
Dropout-5 & \multicolumn{5}{c}{Dropout rate = 0.2} \\
Bilinear-6 & \multicolumn{5}{c}{Resizes the 64$\times$64 input features maps to 127$\times$127}  \\
DecConv-6 & 1$\times$1 & 4$\times$4 & 64 & 2 & ReLU \\
Dropout-6 & \multicolumn{5}{c}{Dropout rate = 0.2} \\
Bilinear-7 & \multicolumn{5}{c}{Resizes the 128$\times$128 input features maps to 64$\times$64}  \\
DecConv-7 & 1$\times$1 & 3$\times$3 & 64 & 1 & ReLU \\
DecConv-8 & 1$\times$1 & 3$\times$3 & 384 & 1 & - \\
\hline
\end{tabular}
\end{center}
\caption{Network architecture of the autoencoder for \ourmethod-S and \ourmethod-M. Layers named ``EncConv'' and ``DecConv'' are standard 2D convolutional layers.}
\label{tab:arch_ae}
\end{table}

\paragraph{Comments on \Cref{alg:training} and \Cref{alg:inference}:}
\begin{itemize}
    \item We use the default initialization method of PyTorch \cite{paszke2019_PyTorch} (version 1.12.0) for the convolutional layers.
    \item We apply the teacher and the student to both the original and the augmented training image. 
That is necessary because the student--teacher model is trained without augmentation, while the autoencoder is trained with augmentation.
During inference, we do not need these second forward passes because images are not augmented at test time.
    \item The sizes of the images of MVTec AD \cite{bergmann2021_mvtec_ad_ijcv, bergmann2019_mvtec_ad_cvpr}, VisA \cite{zou2022spot}, and MVTec LOCO \cite{bergmann2021_mvtec_loco_ijcv} differ.
We resize each input image to $256\times256$ and resize the anomaly map $M$ back to the original image size using bilinear interpolation.
    \item We use a batch size of one.
    \item We use the image normalization of the pretrained models of torchvision \cite{paszke2019_PyTorch}.
That means we subtract $0.485$, $0.456$, and $0.406$ from the R, G, and B channel, respectively, for each input image and divide the channels by $0.229$, $0.224$, and $0.225$, respectively.
We perform this normalization directly before applying a network to an image, i.e., after augmentation.
At test time, this can also be done by adjusting the weights and bias of the first convolutional layer of a network accordingly.
    \item The parameters of the autoencoder $A$ are not only affected by the gradient of $L_\mathrm{AE}$, but also by the gradient of $L_\mathrm{STAE}$.
    \item We obtain an image $P \in \mathbb{R}^{3 \times 256 \times 256}$ from ImageNet by choosing a random image, resizing it to $512\times512$, converting it to gray scale with a probability of $0.3$, and cropping the center $256\times256$ pixels.
\end{itemize}

\subsection{Distillation}
\label{subsec:distillation}

In the following, we describe how to distill the WideResNet-101 \cite{zagoruyko2016wideresnet_wrn} features used by PatchCore \cite{roth2022towards} into the teacher network $T$.
The distillation training algorithm is presented in \Cref{alg:distillation}.
The process in analogous for other pretrained feature extractors.

There are only few requirements regarding the output shape of the feature extractor.
The feature extractors used by PatchCore output features of shape $384\times64\times64$ for an input image size of $512\times512$ pixels.
Therefore, the teacher and the autoencoder also output $384$ channels (as described in \Cref{tab:arch_eads,tab:arch_eadm,tab:arch_ae}).
If a pretrained feature extractor outputs a different number of channels, this default of $384$ output channels of the teacher and the autoencoder can be adjusted flexibly.
During distillation, we resize input images to $512\times512$ for the pretrained feature extractor and to $256\times256$ for the teacher network that is being trained.
This results in an output shape of $384\times64\times64$ for the teacher network as well.
If a feature extractor outputs feature maps of a size other than $64\times64$, we can adjust its input image size to achieve an output feature map size of $64\times64$.
Alternatively, we can adjust the input image size of the teacher network because it is fully convolutional and operates separately on patches of size $33\times33$.
A feature map size of $53\times71$, for example, can be achieved by applying the teacher network to images of size $212\times284$.

We use a batch size of 16 for the distillation training and use ImageNet \cite{russakovsky2015_alexnet} as the pretraining dataset.
We use the official implementation of PatchCore \footnote{\url{https://github.com/amazon-science/patchcore-inspection/tree/6a9a281fc34cb1b13c54b318f71e6f1f371536bb}} and its default values if not stated otherwise.
We use the feature postprocessing of PatchCore as well, which includes pooling features from two layers and projecting each feature vector to a reduced dimensionality of $384$ dimensions, as described in \cite{roth2022towards}.
The features used for our distillation training are the final features used by PatchCore, i.e., those given to the coreset subsampling algorithm when training PatchCore.
We denote the WideResNet-101-based feature extractor, including the feature postprocessing, as $\Psi : \mathbb{R}^{3\times512\times512} \rightarrow \mathbb{R}^{384\times64\times64}$.

\begin{algorithm}
  \caption{Distillation Training Algorithm} 
  \label{alg:distillation} 
  \begin{algorithmic} [1]
    \Require{A pretrained feature extractor $\Psi : \mathbb{R}^{3\times W \times H} \rightarrow \mathbb{R}^{384\times64\times64}$.}
    \Require{A sequence of distillation training images $\mathcal{I}_\mathrm{dist}$}
    \State{Randomly initialize a teacher network $T : \mathbb{R}^{3 \times 256 \times 256} \rightarrow \mathbb{R}^{384 \times 64 \times 64}$ with an architecture as given in \Cref{tab:arch_eads} or \ref{tab:arch_eadm}}
    \For{$c \in {1, \dots, 384}$} \Comment{Compute feature extractor channel normalization parameters $\mu^\Psi \in \mathbb{R}^{384}$ and $\sigma^\Psi \in \mathbb{R}^{384}$}
    \State{Initialize an empty sequence $X \leftarrow (~)$}
    \For{iteration $= 1,2,\dots, \num{10000}$}
        \State{Choose a random training image $I_\mathrm{dist}$ from $\mathcal{I}_\mathrm{dist}$}
        \State{Convert $I_\mathrm{dist}$ to gray scale with a probability of $0.1$}
        \State{Compute $I^\Psi_\mathrm{dist}$ by resizing $I_\mathrm{dist}$ to $3\times W \times H$ using bilinear interpolation}
        \State{$Y^\Psi\leftarrow \Psi(I^\Psi_\mathrm{dist})$}
        \State{$X \leftarrow X^\frown \mathrm{vec}(Y^\Psi_c)$} \Comment{Append the channel output to $X$}
    \EndFor
    \State{Set $\mu^\Psi_c$ to the mean and $\sigma^\Psi_c$ to the standard deviation of the elements of $X$}
    \EndFor
    \State{Initialize the Adam \cite{kingma2014_adam} optimizer with a learning rate of $10^{-4}$ and a weight decay of $10^{-5}$ for the parameters of $T$}
    \For{iteration $= 1,\dots, \num{60000}$}
        \State{$L_\mathrm{batch} \leftarrow 0$}
        \For{batch index $= 1, \dots, 16$}
        \State{Choose a random training image $I_\mathrm{dist}$ from $\mathcal{I}_\mathrm{dist}$}
        \State{Convert $I_\mathrm{dist}$ to gray scale with a probability of $0.1$}
        \State{Compute $I^\Psi_\mathrm{dist}$ by resizing $I_\mathrm{dist}$ to $3\times W \times H$ using bilinear interpolation}
        \State{Compute $I'_\mathrm{dist}$ by resizing $I_\mathrm{dist}$ to $3\times 256 \times 256$ using bilinear interpolation}
        \State{$Y^\Psi\leftarrow \Psi(I^\Psi_\mathrm{dist})$}
        \State{Compute the normalized features $\hat{Y}^\Psi$ given by $\hat{Y}^\Psi_c = (Y^\Psi_c - \mu^\Psi_c) (\sigma^\Psi_c)^{-1}$ for each $c \in \{1, \dots, 384\}$}
        \State{$Y' \leftarrow T(I'_\mathrm{dist})$}
        \State{Compute the squared difference between $\hat{Y}^\Psi$ and $Y'$ for each tuple $(c, w, h)$ as $D^\mathrm{dist}_{c, w, h} = (\hat{Y}^\Psi_{c, w, h} - Y'_{c, w, h})^2$}
        \State{Compute the loss $L_\mathrm{dist}$ as the mean of all elements $D^\mathrm{dist}_{c, w, h}$ of $D^\mathrm{dist}$}
        \State{$L_\mathrm{batch} \leftarrow L_\mathrm{batch} + L_\mathrm{dist}$}
        \EndFor
        \State{$L_\mathrm{batch} \leftarrow 16^{-1} L_\mathrm{batch}$}
        \State{Update the parameters of $T$, denoted by $\theta$, using the gradient $\nabla_\theta L_\mathrm{batch}$ }
    \EndFor
    \State \Return{$T$}
	\end{algorithmic}
\end{algorithm}

\paragraph{Comments on \Cref{alg:distillation}:}
We use the image normalization of the pretrained models of torchvision \cite{paszke2019_PyTorch}.
That means we subtract $0.485$, $0.456$, and $0.406$ from the R, G, and B channel, respectively, for each input image and divide the channels by $0.229$, $0.224$, and $0.225$, respectively.
We perform this normalization directly before applying a network to an image, i.e., after augmentation.

\section{Implementation Details for Other Evaluated Methods}
\label{sec:details_others}

In the following, we provide the implementation and configuration details for Asymmetric Student--Teacher (AST) \cite{rudolph2023asymmetric}, DSR \cite{zavrtanik2022dsr}, FastFlow \cite{yu2021fastflow}, GCAD \cite{bergmann2021_mvtec_loco_ijcv}, PatchCore \cite{roth2022towards}, SimpleNet \cite{Liu_2023_CVPR}, and Student--Teacher \cite{bergmann2020_uninformed_cvpr}.

\subsection{AST}
\label{sec:details_ast}

We use the official implementation of Rudolph \etal \cite{rudolph2023asymmetric} \footnote{\url{https://github.com/marco-rudolph/AST/tree/1a157973a0e2cb23b6fbb853db8ae43537ab2568}}.
We use the default configuration without modifications, but are not able to fully reproduce the results reported in the AST paper.
The AST paper reports a mean image-level detection AU-ROC of \SI{99.2}{\percent} on MVTec AD, averaged across five runs.
We obtain an AU-ROC of \SI{98.9}{\percent} across five runs.

\subsection{DSR}
\label{sec:details_dsr}

We use the official implementation of Zavrtanik \etal \cite{zavrtanik2022dsr} \footnote{\url{https://github.com/VitjanZ/DSR_anomaly_detection/tree/672bfb81434fd2a6c5ef00db858cef8834c54f28}}.
We use the default configuration without modifications for reproducing the results on MVTec AD.
We obtain a mean image-level detection AU-ROC of \SI{98.1}{\percent} on MVTec AD, which is close to the \SI{98.2}{\percent} reported by the authors.
On the scenarios from VisA, which contain more training images than those of MVTec AD, we change the number of epochs to 50 to keep the total number of training iterations in a similar range.

\subsection{FastFlow}
\label{sec:details_fastflow}

We use the implementation of Akcay \etal \cite{akcay2022anomalib} \footnote{\url{https://github.com/openvinotoolkit/anomalib/tree/e66a17c86489486f6bbd5099366383e8660923fd}}.
We use the FastFlow version based on the WideResNet-50-2 feature extractor, as it is similar to the WideResNet used by PatchCore, SimpleNet and our method.
We use the default configuration, but disable early stopping, i.e., the scenario-specific tuning of the training duration on test images.
Instead, we choose a constant training duration (200 steps) that works well on average for all evaluated datasets.

\subsection{GCAD}
\label{sec:details_gcad}

We implement GCAD as described by Bergmann \etal \cite{bergmann2021_mvtec_loco_ijcv}.
We are able to reproduce the results reported by the authors, but adapt GCAD to a configuration that performs better in our experiments.
GCAD consists of an ensemble of two anomaly detection models that use different feature extractors.
The first member uses a feature extractor that operates on patches of size $17\times17$ while the feature extractor used by the second member operates on patches of size $33\times33$.
We find that the second member performs better on average than the combined ensemble and therefore report the results for this member in the main paper.
On the logical anomalies of MVTec LOCO, the single model scores an image-level detection AU-ROC of \SI{83.9}{\percent}, while the AU-ROC of the ensemble model used by the authors is \SI{86.0}{\percent}.
The overall anomaly detection performance on MVTec LOCO, however, stays the same.

\subsection{PatchCore}
\label{sec:details_patchcore}

We use the official implementation of Roth \etal \cite{roth2022towards} \footnote{\url{https://github.com/amazon-science/patchcore-inspection/tree/6a9a281fc34cb1b13c54b318f71e6f1f371536bb}} and are able to reproduce the results reported for MVTec AD.
As described in the main paper, we disable the cropping of the center \SI{76.6}{\percent} of input images for a fair comparison.
For the single model variant of PatchCore, we use the configuration of PatchCore for which the authors report the lowest latency.
Specifically, this means setting the coreset subsampling ratio to \SI{1}{\percent}, the image size to $224\times224$ pixels, and the feature extraction backbone to a WideResNet-101.
For the ensemble variant, we use the configuration for which the authors report the best detection AU-ROC on MVTec AD.
We use a WideResNet-101, a ResNeXT-101 \cite{xie2017resnext}, and a DenseNet-201 \cite{huang2017densely} as backbones, set the coreset subsampling ratio to \SI{1}{\percent}, and use images of size $320\times320$ pixels.

\subsection{SimpleNet}
\label{sec:details_simplenet}
We use the official implementation of Liu \etal \cite{Liu_2023_CVPR} \footnote{\url{https://github.com/DonaldRR/SimpleNet/tree/35bf32292995842a4277a7c93431430129efccb5}} and are able to reproduce the reported results.
As explained in the main paper, we disable the scenario-specific tuning of the training duration on test images for a fair comparison.

\subsection{Student--Teacher}
\label{sec:details_st}

We implement the original multi-scale Student--Teacher (\studteach) method as described by Bergmann \etal \cite{bergmann2020_uninformed_cvpr}.
We use the default hyperparameter settings without modification.
Our implementation achieves better anomaly localization results on MVTec AD than those reported by the authors but matches those reported in \cite{bergmann2021_mvtec_ad_ijcv}.

\section{Robustness to the Distillation Backbone Architecture}
\label{sec:backbones}
In the main paper, we use the features from a WideResNet-101 for training a teacher network in \Cref{alg:distillation}.
The default configuration of PatchCore uses the same features.
In \Cref{tab:distillation_backbones}, we evaluate the anomaly detection performance for other backbones.
Specifically, we evaluate the two additional backbones that \patchcoreens\ uses, i.e., a ResNeXt-101 and a DenseNet-201.
On the three evaluated dataset collections, the anomaly detection performance of \ourmethod\ is similarly robust to the choice of the backbone in comparison to the robustness of PatchCore.
On MVTec AD, both methods perform very similarly across backbones, while their performance on MVTec LOCO varies more.
On VisA, the gap between the structural anomaly detection performance of PatchCore and that of \ourmethod\ becomes evident.

\begin{table}[h]
\begin{center}
\begin{tabular}{ccccc}
& Method & WideResNet-101 & ResNeXt-101 & DenseNet-201 \\
\hline
\multirow{3}{*}{\rotatebox{90}{\specialcell[c]{MVTec \\ AD}}} & \text{PatchCore} & 98.7 & 98.8 & 98.7 \\
& \ourmethod-S & 98.8 & 98.9 & 98.8 \\
& \ourmethod-M & 99.1 & 99.0 & 99.2 \\
&&&& \\ 
\multirow{3}{*}{\rotatebox{90}{\specialcell[c]{MVTec \\ LOCO}}} & \text{PatchCore} & 80.3  & 78.9 & 76.5 \\
& \ourmethod-S & 90.0 & 90.1 & 90.6 \\
& \ourmethod-M & 90.7 & 89.9 & 88.3 \\
&&&& \\ 
\multirow{3}{*}{\rotatebox{90}{VisA}} & \text{PatchCore} & 94.3 & 95.2 & 94.8 \\
& \ourmethod-S & 97.5 & 97.3 & 97.1 \\
& \ourmethod-M & 98.1 & 98.0 & 97.7 \\
\end{tabular}
\end{center}
\caption{Mean anomaly detection AU-ROC percentages for different backbones. For \ourmethod, each listed architecture is used as the distillation backbone in \Cref{alg:distillation}. The ``WideResNet-101'' column contains the results reported in \Cref{tab:per_dataset} in the main paper.}
\label{tab:distillation_backbones}
\end{table}

\clearpage

\section{Additional Anomaly Detection Metrics}
\label{sec:ad_metrics}

In this section, we report the results for additional anomaly detection metrics.
\Cref{sec:ad_metrics_classification} evaluates image-level anomaly detection metrics.
\Cref{sec:ad_metrics_localization} evaluates pixel-level anomaly localization metrics.

For per-scenario evaluation results, see the \href{https://www.mydrive.ch/shares/79401/c41dcdb937972fb43d5cdd7bfa7072f8/download/449203483-1690527455/per_scenario_results.json}{\texttt{per\_scenario\_results.json}} file \footnote{\url{https://www.mydrive.ch/shares/79401/c41dcdb937972fb43d5cdd7bfa7072f8/download/449203483-1690527455/per_scenario_results.json}}.

Following the official MVTec LOCO evaluation script \footnote{\url{https://www.mvtec.com/company/research/datasets/mvtec-loco}}, we evaluate each performance metric separately on the structural and on the logical anomalies of MVTec LOCO.
Then, we compute the mean of the two scores to compute the overall performance of a method on a scenario of MVTec LOCO.

\subsection{Anomaly Detection}
\label{sec:ad_metrics_classification}

In the main paper, we evaluate the image-level anomaly detection performance with the area under the ROC curve (AU-ROC).
Here, we report the results for the area under the precision recall curve (AU-PRC) as well.
For information on the differences between the AU-ROC and the AU-PRC, we refer to Davis and Goadrich \cite{davis2006relationship}.

\Cref{tab:classification_roc} shows the anomaly detection performance of each method measured with the AU-ROC.
This table contains the results reported in the main paper.
\Cref{tab:classification_prc} shows the results for the image-level AU-PRC.

\begin{table}[h!]
\begin{center}
\begin{tabular}{c|c|c|ccc||c}
Method & \specialcell[c]{MAD \\ Mean} & \specialcell[c]{VisA \\ Mean} & \specialcell[c]{LOCO \\ Structural} & \specialcell[c]{LOCO \\ Logical} & \specialcell[c]{LOCO \\ Mean} & \specialcell[c]{Overall \\ Mean} \\
\hline
GCAD & 89.1 & 83.7 & 82.7 & 83.9 & 83.3 & 85.4 \\
SimpleNet & 98.2 & 87.9 & 83.7 & 71.5 & 77.6 & 87.9 \\
\studteach & 93.2 & 94.6 & 88.3 & 66.5 & 77.4 & 88.4 \\
FastFlow & 96.9 & 93.9 & 82.9 & 75.5 & 79.2 & 90.0 \\
DSR & 98.1 & 91.8 & 90.2 & 75.0 & 82.6 & 90.8 \\
PatchCore & 98.7 & 94.3 & 84.8 & 75.8 & 80.3 & 91.1 \\
\patchcoreens & \textbf{99.3} & 97.7 & 87.7 & 71.0 & 79.4 & 92.1 \\
AST & 98.9 & 94.9 & 87.1 & 79.7 & 83.4 & 92.4 \\
\hline
\ourmethod-S & 98.8 & 97.5 & 94.1 & 85.8 & 90.0 & 95.4 \\
\ourmethod-M & 99.1 & \textbf{98.1} & \textbf{94.7} & \textbf{86.8} & \textbf{90.7} & \textbf{96.0} \\
\end{tabular}
\end{center}
\caption{Mean anomaly detection AU-ROC percentages per dataset collection. For \ourmethod, we report the mean of five runs.}
\label{tab:classification_roc}
\vspace*{-5mm}
\end{table}

\begin{table}[h!]
\begin{center}
\begin{tabular}{c|c|c|ccc||c}
Method & \specialcell[c]{MAD \\ Mean} & \specialcell[c]{VisA \\ Mean} & \specialcell[c]{LOCO \\ Structural} & \specialcell[c]{LOCO \\ Logical} & \specialcell[c]{LOCO \\ Mean} & \specialcell[c]{Overall \\ Mean} \\
\hline
GCAD & 95.7 & 87.1 & 81.0 & 84.9 & 83.0 & 88.6 \\
SimpleNet & 98.5 & 90.1 & 82.5 & 73.5 & 78.0 & 88.9 \\
\studteach & 95.7 & 94.6 & 87.9 & 70.7 & 79.3 & 89.9 \\
FastFlow & 95.3 & 94.7 & 79.5 & 76.2 & 77.9 & 89.3 \\
DSR & 98.1 & 93.8 & 88.2 & 76.6 & 82.4 & 91.4 \\
PatchCore & 98.9 & 95.2 & 84.6 & 77.7 & 81.2 & 91.8 \\
\patchcoreens & \textbf{99.0} & 97.8 & 88.3 & 74.7 & 81.5 & 92.8 \\
AST & 98.9 & 95.3 & 84.5 & 80.5 & 82.5 & 92.2 \\
\hline
\ourmethod-S & 98.7 & 97.5 & 93.6 & 86.2 & 89.9 & 95.4 \\
\ourmethod-M & 98.9 & \textbf{98.0} & \textbf{93.9} & \textbf{86.8} & \textbf{90.3} & \textbf{95.7} \\
\end{tabular}
\end{center}
\caption{Mean anomaly detection AU-PRC percentages per dataset collection. For \ourmethod, we report the mean of five runs.}
\label{tab:classification_prc}
\vspace*{-5mm}
\end{table}

\subsection{Anomaly Localization}
\label{sec:ad_metrics_localization}

To evaluate the anomaly localization performance, we use the area under the PRO curve (AU-PRO) up to a false positive rate (FPR) of \SI{30}{\percent} in the main paper, as recommended by \cite{bergmann2021_mvtec_ad_ijcv}.
The AU-PRO metric \cite{bergmann2021_mvtec_ad_ijcv} is similar to the pixel-wise AU-ROC.
The difference is that the pixel-wise AU-ROC gives each ground truth defect \textit{pixel} the same weight in its computation.
The AU-PRO gives each ground truth defect \textit{region} the same weight.
The FPR limit of \SI{30}{\percent} is due to the fact that a method that segments, on average, more than \SI{30}{\percent} of defect-free pixels as anomalous is of limited use.

\Cref{tab:localization_pro_03} contains the results reported in the main paper.
Here, we report the AU-PRO for an FPR limit of \SI{5}{\percent} as well in \Cref{tab:localization_pro_005}.
For comparison, we also report the pixel-wise AU-ROC for an FPR limit of \SI{5}{\percent} in \Cref{tab:localization_roc_005}.
Furthermore, we evaluate the pixel-wise AU-PRC as an additional segmentation, and thus, localization performance metric in \Cref{tab:localization_prc}.
The \href{https://www.mydrive.ch/shares/79401/c41dcdb937972fb43d5cdd7bfa7072f8/download/449203483-1690527455/per_scenario_results.json}{\texttt{per\_scenario\_results.json}} file \footnote{\url{https://www.mydrive.ch/shares/79401/c41dcdb937972fb43d5cdd7bfa7072f8/download/449203483-1690527455/per_scenario_results.json}} also contains the AU-PRO and pixel-wise AU-ROC results for an FPR limit of \SI{100}{\percent}.

\begin{table}[h!]
\begin{center}
\begin{tabular}{c|c|c|ccc||c}
Method & \specialcell[c]{MAD \\ Mean} & \specialcell[c]{VisA \\ Mean} & \specialcell[c]{LOCO \\ Structural} & \specialcell[c]{LOCO \\ Logical} & \specialcell[c]{LOCO \\ Mean} & \specialcell[c]{Overall \\ Mean} \\
\hline
GCAD & 91.0 & 83.7 & 89.5 & 89.4 & 89.5 & 88.0 \\
SimpleNet & 89.6 & 68.9 & 60.6 & 68.6 & 64.6 & 74.4 \\
\studteach & 92.4 & 93.0 & 90.8 & 76.4 & 83.6 & 89.7 \\
FastFlow & 92.5 & 86.8 & 84.2 & 76.5 & 80.3 & 86.5 \\
DSR & 90.8 & 68.1 & 81.3 & 72.3 & 76.8 & 78.6 \\
PatchCore & 92.7 & 79.7 & 64.3 & 76.6 & 70.4 & 80.9 \\
\patchcoreens & \textbf{95.6} & 79.3 & 62.0 & 72.6 & 67.3 & 80.7 \\
AST & 81.2 & 81.5 & 75.4 & 62.6 & 69.0 & 77.2 \\
\hline
\ourmethod-S & 93.1 & 93.1 & 92.6 & 90.1 & 91.3 & 92.5 \\
\ourmethod-M & 93.5 & \textbf{94.0} & \textbf{93.7} & \textbf{91.3} & \textbf{92.5} & \textbf{93.3} \\
\end{tabular}
\end{center}
\caption{Mean anomaly localization performance per method and dataset collection, measured with the AU-PRO up to a FPR of \SI{30}{\percent}. For \ourmethod, we report the mean of five runs.}
\label{tab:localization_pro_03}
\vspace*{-5mm}
\end{table}

\begin{table}[h!]
\begin{center}
\begin{tabular}{c|c|c|ccc||c}
Method & \specialcell[c]{MAD \\ Mean} & \specialcell[c]{VisA \\ Mean} & \specialcell[c]{LOCO \\ Structural} & \specialcell[c]{LOCO \\ Logical} & \specialcell[c]{LOCO \\ Mean} & \specialcell[c]{Overall \\ Mean} \\
\hline
GCAD & 68.8 & 52.6 & 68.8 & 67.1 & 68.0 & 63.1 \\
SimpleNet & 61.8 & 37.7 & 36.6 & 36.1 & 36.3 & 45.3 \\
\studteach & 73.4 & 75.0 & 75.6 & 49.7 & 62.6 & 70.4 \\
FastFlow & 71.6 & 63.4 & 64.5 & 49.1 & 56.8 & 63.9 \\
DSR & 78.9 & 49.5 & 67.1 & 49.8 & 58.5 & 62.3 \\
PatchCore & 68.6 & 49.4 & 37.9 & 41.5 & 39.7 & 52.6 \\
\patchcoreens & \textbf{79.5} & 55.1 & 37.8 & 35.3 & 36.5 & 57.1 \\
AST & 42.1 & 48.0 & 50.1 & 35.3 & 42.7 & 44.3 \\
\hline
\ourmethod-S & 78.2 & 73.4 & 80.8 & 74.8 & 77.8 & 76.5 \\
\ourmethod-M & 78.4 & \textbf{75.9} & \textbf{83.2} & \textbf{76.5} & \textbf{79.8} & \textbf{78.0} \\
\end{tabular}
\end{center}
\caption{Mean anomaly localization performance per method and dataset collection, measured with the AU-PRO up to a FPR of \SI{5}{\percent}. For \ourmethod, we report the mean of five runs.}
\label{tab:localization_pro_005}
\vspace*{-5mm}
\end{table}

\begin{table}[h!]
\begin{center}
\begin{tabular}{c|c|c|ccc||c}
Method & \specialcell[c]{MAD \\ Mean} & \specialcell[c]{VisA \\ Mean} & \specialcell[c]{LOCO \\ Structural} & \specialcell[c]{LOCO \\ Logical} & \specialcell[c]{LOCO \\ Mean} & \specialcell[c]{Overall \\ Mean} \\
\hline
GCAD & 72.1 & 73.1 & 73.1 & 32.0 & 52.5 & 65.9 \\
SimpleNet & 67.9 & 57.1 & 36.4 & 22.1 & 29.2 & 51.4 \\
\studteach & 74.3 & 82.7 & 69.8 & 20.6 & 45.2 & 67.4 \\
FastFlow & 72.1 & 78.9 & 63.4 & 33.7 & 48.6 & 66.5 \\
DSR & 76.1 & 66.5 & 66.0 & 25.5 & 45.7 & 62.8 \\
PatchCore & 74.1 & 65.0 & 43.5 & 24.1 & 33.8 & 57.6 \\
\patchcoreens & 79.4 & 65.7 & 38.7 & 20.4 & 29.6 & 58.2 \\
AST & 41.1 & 67.4 & 52.4 & 30.9 & 41.7 & 50.1 \\
\hline
\ourmethod-S & \textbf{79.7} & 86.3 & 80.6 & 33.8 & 57.2 & 74.4 \\
\ourmethod-M & 79.4 & \textbf{86.9} & \textbf{82.1} & \textbf{35.3} & \textbf{58.7} & \textbf{75.0} \\
\end{tabular}
\end{center}
\caption{Mean anomaly localization performance per method and dataset collection, measured with the AU-ROC up to a FPR of \SI{5}{\percent}. For \ourmethod, we report the mean of five runs.}
\label{tab:localization_roc_005}
\vspace*{-5mm}
\end{table}

\begin{table}[h!]
\begin{center}
\begin{tabular}{c|c|c|ccc||c}
Method & \specialcell[c]{MAD \\ Mean} & \specialcell[c]{VisA \\ Mean} & \specialcell[c]{LOCO \\ Structural} & \specialcell[c]{LOCO \\ Logical} & \specialcell[c]{LOCO \\ Mean} & \specialcell[c]{Overall \\ Mean} \\
\hline
GCAD & 59.3 & 27.8 & 41.4 & 38.7 & 40.1 & 42.4 \\
SimpleNet & 51.5 & 22.6 & 11.9 & 29.3 & 20.6 & 31.6 \\
\studteach & 59.9 & 36.2 & 43.5 & 27.4 & 35.4 & 43.8 \\
FastFlow & 57.6 & 33.4 & 35.1 & 41.3 & 38.2 & 43.1 \\
DSR & \textbf{69.2} & \textbf{41.1} & 50.4 & 32.7 & 41.5 & 50.6 \\
PatchCore & 57.6 & 27.8 & 17.8 & 32.5 & 25.2 & 36.8 \\
\patchcoreens & 64.1 & 28.3 & 15.1 & 28.9 & 22.0 & 38.2 \\
AST & 29.7 & 22.9 & 17.0 & 35.6 & 26.3 & 26.3 \\
\hline
\ourmethod-S & 65.9 & 40.4 & \textbf{54.0} & 40.2 & \textbf{47.1} & \textbf{51.1} \\
\ourmethod-M & 63.8 & 40.8 & 51.9 & \textbf{42.0} & 46.9 & 50.5 \\
\end{tabular}
\end{center}
\caption{Mean anomaly localization performance per method and dataset collection, measured with the pixel-wise AU-PRC.  For \ourmethod, we report the mean of five runs.}
\label{tab:localization_prc}
\end{table}

\section{Timing Methodology and Additional Computational Efficiency Metrics}
\label{sec:efficiency_metrics}

\begin{table}[h!]
\begin{center}
\begin{tabular}{cccccccc}
Method & \specialcell[c]{Detect. \\ AU-ROC} & \specialcell[c]{Segment. \\ AU-PRO} & \specialcell[c]{Latency \\ {[ms]}} & \specialcell[c]{Throughput \\ {[img / s]}} & \specialcell[c]{Number of \\ Parameters [$\times 10^6$]} & \specialcell[c]{FLOPs \\ {[$\times 10^9$]}} &  \specialcell[c]{GPU Memory \\  {[MB]}} \\
\hline
GCAD & 85.4 & 88.0 & 11 & 121 & 65 & 416 & 555 \\
SimpleNet & 87.9 & 74.4 & 12 & 194 & 73 & \textbf{38} & 508 \\
\studteach & 88.4 & 89.7 & 75 & 16 & 26 & 4468 & 1077 \\
FastFlow & 90.0 & 86.5 & 17 & 120 & 92 & 85 & 404 \\
DSR & 90.8 & 78.6 & 17 & 104 & 40 & 267 & 314 \\
PatchCore & 91.1 & 80.9 & 32 & 76 & 83 + 3 & 41 + kNN & 637 + kNN \\
\patchcoreens & 92.1 & 80.7 & 148 & 13 & 150 + 8 & 159 + kNN & 1335 + kNN \\
AST & 92.4 & 77.2 & 53 & 41 & 154 & 199 & 618 \\
\hline
\ourmethod-S & \meanwithstd{95.4}{0.06} & \meanwithstd{92.5}{0.05} & \meanwithstd{\textbf{2.2}}{0.01} & \meanwithstd{\textbf{614}}{2} & \meanwithstd{\textbf{8}}{0} & \meanwithstd{76}{0} & \meanwithstd{\textbf{100}}{0} \\
\ourmethod-M & \meanwithstd{\textbf{96.0}}{0.09} & \meanwithstd{\textbf{93.3}}{0.04} & \meanwithstd{4.5}{0.01} & \meanwithstd{269}{1} & \meanwithstd{21}{0} & \meanwithstd{235}{0} & \meanwithstd{161}{0} \\
\end{tabular}
\end{center}
\caption{
Extension of \Cref{tab:main} in the main paper by additional computational efficiency metrics measured on an NVIDIA RTX A6000 GPU.
For \ourmethod, we report the mean and standard deviation of five runs.
For PatchCore, we report the computational requirements of the feature extraction during inference separately from the nearest neighbor search.
}
\label{tab:efficiency}
\end{table}

In the following, we describe how we measure the latency and the throughput of each anomaly detection method.
Latency refers to the inference runtime, i.e., how long it takes a method to generate the anomaly detection result for a single image.
Throughput refers to how many images can be processed per second when allowing a batched processing of images.
In settings in which latency constraints are fulfilled or not present, a high throughput is relevant for using computational resources efficiently and thus for reducing the economic cost of an application.

All evaluated methods are implemented in PyTorch.
All of them, including the nearest neighbor search of PatchCore, run faster on each of the GPUs in our experimental setup than on a CPU.
We therefore execute each method on a GPU.
For a test image, our timing begins with the transfer of the image from the CPU to the GPU.
We include the transfer to regard the benefit of a method that would run exclusively on a CPU.
Our timing stops when the anomaly detection result, which for all evaluated methods is an anomaly map, is available on the CPU.
For each method, we remove unnecessary parts for the timing, such as the computation of losses during inference, and use float16 precision for all networks.
Switching from float32 to float16 for the inference of \ourmethod\ does not change the anomaly detection results for the 32 anomaly detection scenarios evaluated in this paper.
In latency-critical applications, padding in the PDN architecture of \ourmethod\ can be disabled.
This speeds up the forward pass of the PDN architecture by \SI{80}{\micro\second} without impairing the detection of anomalies.
We time \ourmethod\ without padding and therefore report the anomaly detection results for this setting in the experimental results of this paper.
We perform 1000 forward passes as warm up and report the mean runtime of the following 1000 forward passes.
For the latency, we report the average runtime of 1000 forward passes with a batch size of 1.
We compute the throughput by dividing \num{16000} by the sum of the runtimes of 1000 forward passes with a batch size of 16.
In addition to the latency and the throughput, we report the number of parameters, the number of floating point operations (FLOPs), and the GPU memory consumption for each method in \Cref{tab:efficiency}.
Analogously to the latency, we measure these metrics for the processing of one image during inference and report the mean of 1000 forward passes.
The number of parameters and the FLOPs remain constant across forward passes, while the GPU memory consumption varies slightly (less than one MB difference between forward passes).

\paragraph{Technical Details}
For methods that use features from hidden layers of a pretrained network, we exclude the layers that are not required for computing these features, i.e., classification heads etc.
We measure the number of FLOPs using the official profiling framework of PyTorch \cite{paszke2019_PyTorch} (version 1.12.0).
Specifically, we wrap the inference function of a method into a call of \texttt{with torch.profiler.profile(with\_flops=True) as prof:}.
For measuring the GPU memory consumption, we also use the official profiling framework of PyTorch.
We obtain the peak of the reserved GPU memory during inference with \texttt{torch.cuda.memory\_stats()['reserved\_bytes.all.peak']}.

\paragraph{Interpretability of Efficiency Metrics}
In the main paper, we focus on the latency and the throughput of the evaluated anomaly detection methods.
The number of parameters and the number of FLOPs are often used as proxy metrics for the runtime, but can be misleading.
For example, the number of parameters of FastFlow in \Cref{tab:efficiency} is roughly 2.5 times larger than that of S--T.
Yet, the latency of FastFlow is substantially lower and its throughput is 6.5 times higher.

With 4.5 trillion FLOPs, S--T exceeds the FLOPs of other methods by a large margin.
The high number of FLOPs, however, comes from the fact that S--T uses convolutions that operate on large feature maps.
This means that these convolutions can be parallelized well on a GPU, while implementing them naively on a CPU would indeed cause a prohibitively long runtime.
FLOPs measurements do not account for this, because they do not consider how well operations can be parallelized.
The number of FLOPs can therefore be an unreliable metric for efficiency.
For example, the number of FLOPs of S--T is more than \SI{2000}{\percent} higher than that of AST, but the latency is only \SI{42}{\percent} higher.

The GPU memory footprint of a method can theoretically be reduced drastically by freeing obsolete GPU memory segments after each layer's execution during a forward pass.
In the extreme case, one could even directly free the memory of individual input activation values directly after the output activation of a neuron in a convolutional layer has been computed.
This, however, would worsen the runtime of a forward pass, which generally improves when reserved GPU memory segments can be reused.
Therefore, the GPU memory footprint of a method needs to be reported and analyzed jointly with the latency and throughput.
We focus on the GPU memory required for achieving the reported latency and throughput and therefore measure the peak of the reserved GPU memory during a forward pass.

\paragraph{PatchCore}
For PatchCore, we distinguish between the backbones used to compute features and the kNN algorithm itself.
For example, the part of the WideResNet-101 backbone until the layer used for computing features has 83 million parameters.
During training, PatchCore computes the feature vectors of all training images.
The coreset subsampling phase of PatchCore reduces the number of feature vectors to \SI{1}{\percent} of the computed feature vectors.
These are then indexed and stored in GPU memory to enable a fast search for nearest neighbors during inference.
This, however, means that the number of parameters, the FLOPs, and the GPU memory footprint of PatchCore depend on the training images.
We therefore benchmark PatchCore on the ``cashew'' scenario of VisA, which contains 450 training images and is thus closest to the average 439 training images of the 32 scenarios of MVTec AD, VisA, and MVTec LOCO.
We do not report the FLOPs and the GPU memory consumption of the kNN search, as we were not able to measure it with the kNN library used by the official PatchCore implementation.
The number of parameters of the kNN search is given by the number of values stored in the GPU memory during inference.
In the case of \patchcoreens, for example, the search database contains 8 million values.

\clearpage

\begin{figure}[h!]
\hfill
\subfigure[Anomaly detection]{\includegraphics[width=.49\linewidth]{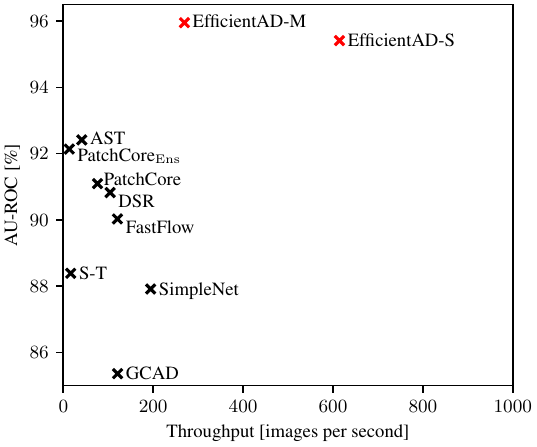}}
\hfill
\subfigure[Anomaly localization]{\includegraphics[width=.49\linewidth]{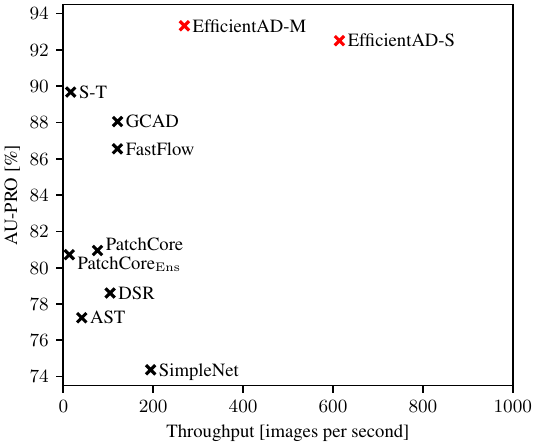}}
\hfill
\caption{Anomaly detection performance vs. throughput on an NVIDIA RTX A6000 GPU\@. We report the image-level anomaly detection performance on the left using the image-level AU-ROC. On the right, we report the anomaly localization performance using the pixel-level AU-PRO segmentation metric up to a FPR of \SI{30}{\percent}. Each AU-ROC and AU-PRO value is an average of the values on MVTec AD, VisA, and MVTec LOCO. We measure the throughput using a batch size of 16.
}
\label{fig:throughput}
\end{figure}

\paragraph{Anomaly Detection and Throughput}
\Cref{fig:throughput} shows the anomaly detection performance together with the throughput of each evaluated method, analogous to \Cref{fig:teaser} in the main paper. 
Apart from \ourmethod, the ranking of methods changes drastically between the image-level and the pixel-level detection of anomalies.

\paragraph{Latency per GPU}
In \Cref{tab:latency_per_gpu}, we provide the values for \Cref{fig:latency_per_gpu} in the main paper.

\begin{table}[h!]
\begin{center}
\begin{tabular}{cccccc}
Method & RTX A6000 & RTX A5000 & Tesla V100 & RTX 3080 & RTX 2080 Ti \\
\hline
\ourmethod-S & \textbf{2.2} & \textbf{2.5} & \textbf{3.9} & \textbf{3.8} & \textbf{4.5} \\
\ourmethod-M & 4.5 & 5.3 & 6.3 & 7.0 & 7.6 \\
GCAD & 10.7 & 11.7 & 12.9 & 13.7 & 18.0 \\
SimpleNet & 12.0 & 13.3 & 19.2 & 18.1 & 21.9 \\
FastFlow & 16.5 & 17.1 & 26.1 & 27.5 & 31.0 \\
DSR & 17.2 & 18.0 & 24.8 & 24.6 & 34.5 \\
PatchCore & 32.0 & 31.5 & 47.1 & 41.1 & 53.2 \\
AST & 53.1 & 53.4 & 75.6 & 82.3 & 87.1 \\
\studteach & 74.7 & 81.0 & 82.2 & 99.6 & 121.7 \\
\patchcoreens & 147.6 & 145.0 & 229.2 & 189.0 & 216.9 \\
\end{tabular}
\end{center}
\caption{
Latency in milliseconds per GPU, as plotted in \Cref{fig:latency_per_gpu} in the main paper.
}
\label{tab:latency_per_gpu}
\end{table}

\clearpage

\section{Qualitative Results}
\label{sec:qualitative_results}

\begin{figure}
\begin{center}
\includegraphics[width=1.0\linewidth]{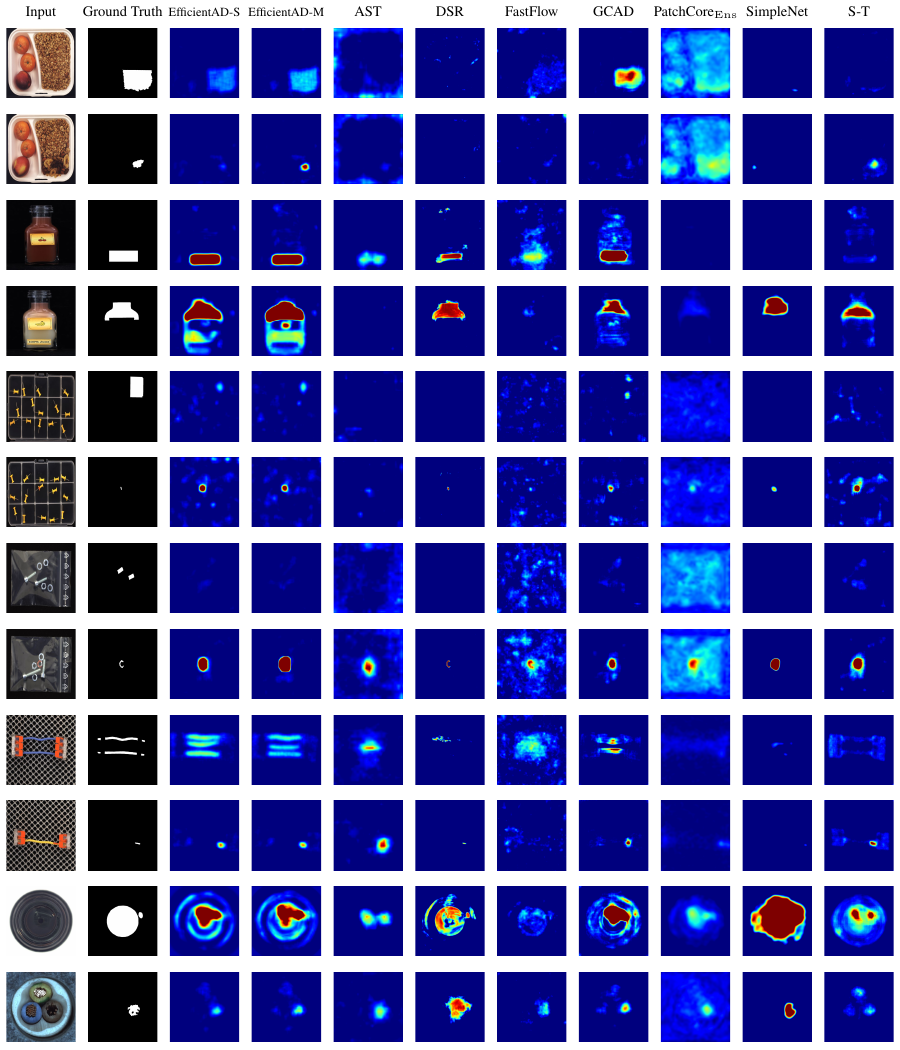}
\end{center}
\caption{
Anomaly maps on anomalous images from MVTec LOCO and MVTec AD.
For MVTec LOCO, we show a logical anomaly (upper row) and a structural anomaly (lower row) for each scenario.
The receptive field of AST's features is large enough to detect some logical anomalies, while \patchcoreens\ and S--T struggle with logical anomalies.
}
\label{fig:qualitative_1}
\end{figure}

\begin{figure}
\begin{center}
\includegraphics[width=1.0\linewidth]{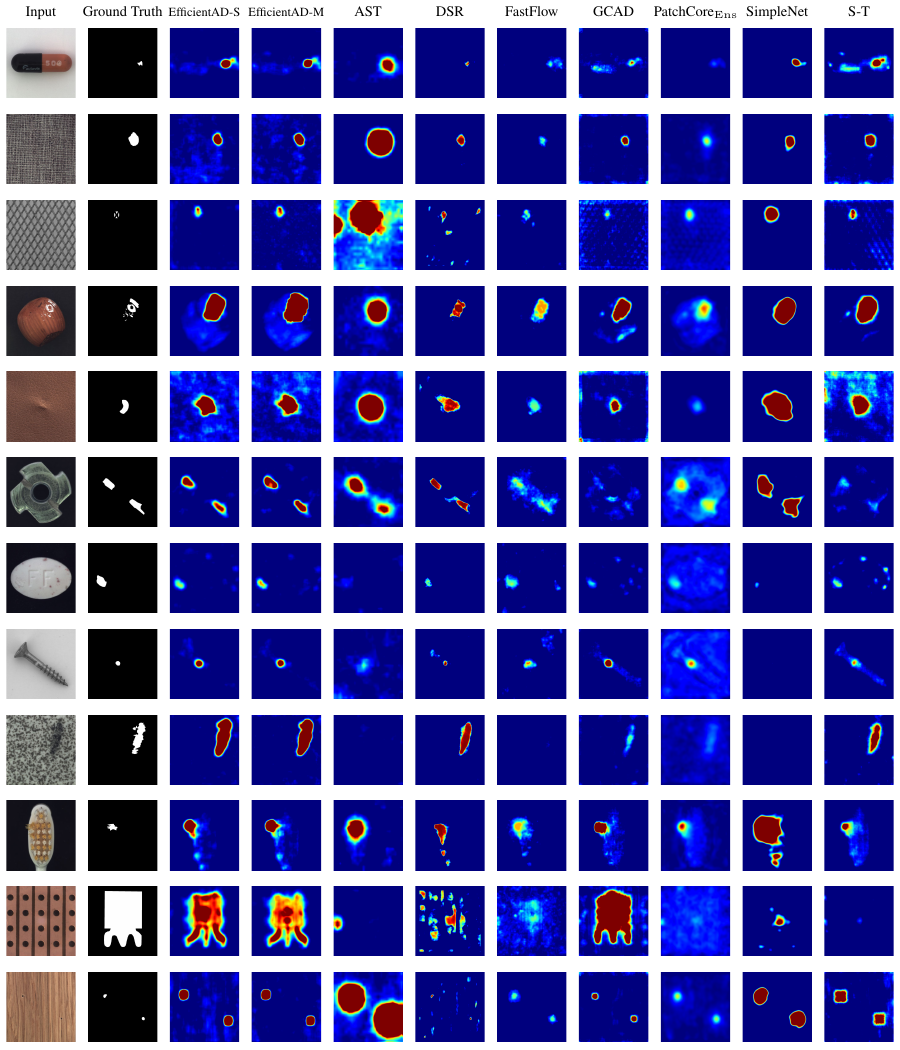}
\end{center}
\caption{
Anomaly maps on anomalous images from MVTec AD.
Almost all anomalies are detected by every method, but the separability of pixel anomaly scores varies between methods.
For example, \patchcoreens\ detects the anomaly on the capsule in the first row but the pixel anomaly scores are in a similar range as the false positive detections in the background of the screw image.
}
\label{fig:qualitative_2}
\end{figure}

\begin{figure}
\begin{center}
\includegraphics[width=1.0\linewidth]{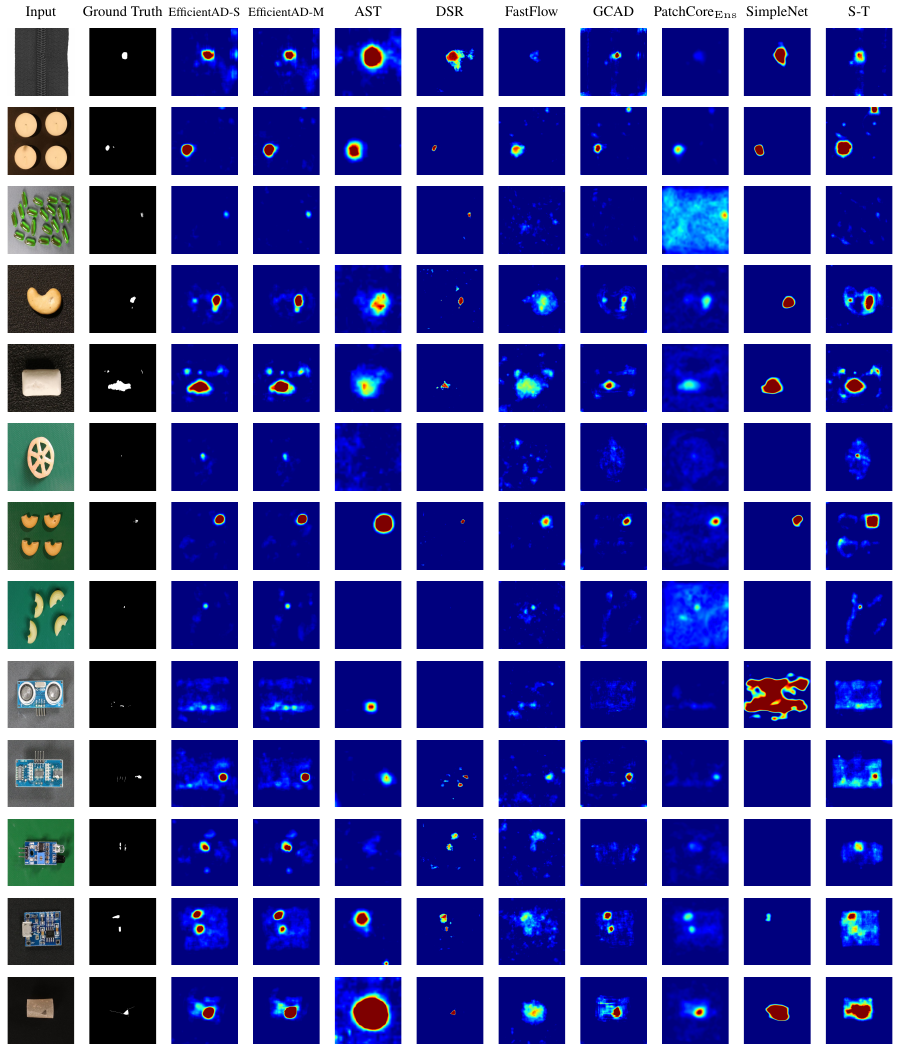}
\end{center}
\caption{
Anomaly maps on anomalous images from MVTec AD and VisA.
VisA contains challenging, small anomalies, such as the defect on the non-aligned macaronis or the defect on the fryum two rows above.
}
\label{fig:qualitative_3}
\end{figure}

In \Cref{fig:qualitative_1,fig:qualitative_2,fig:qualitative_3}, we display anomaly maps for each of the 32 scenarios of MVTec AD, VisA, and MVTec LOCO.
For MVTec LOCO, we show both logical and structural anomalies.
We visualize the anomaly maps using a different scale for each method, since the anomaly score scales differ between methods.
Across scenarios, however, we use the same color scale per anomaly detection method.
A consistent anomaly score scale across applications is an important requirement for a method.
Otherwise, the scale of scores on anomalies is hard to forecast if no or only few defect images are present during the development of the anomaly detection system.
Knowing the scale is important for choosing a robust threshold value that ultimately determines whether an image or a pixel is anomalous or not.
Furthermore, a consistent scale facilitates the interpretation of anomaly maps.
For the evaluated methods, we choose the start and end values of the color scales so that true positive and true negative detections become clearly visible.
For example, the color scale of AST ranges from 2 to 10.
Scores outside of this range are visualized with the minimum and maximum color value, respectively.
For PatchCore, choosing the range of the color scale is difficult.
On the one hand, scores of true positive detections are low, such as the contamination of the banana juice bottle in \Cref{fig:qualitative_1}.
On the other hand, scores of false positive detections are similarly high, such as the predictions on the breakfast box in \Cref{fig:qualitative_1}.

Overall, the evaluated anomaly detection methods succeed on the anomalies of MVTec AD, but leave room for improvement on MVTec LOCO and VisA.
\begin{itemize}
\item EfficientAD responds to both logical and structural anomalies in the images.
The strength of its response sometimes leaves room for improvement, for example, on the logical anomalies of the breakfast box and the box of pushpins in \Cref{fig:qualitative_1}.
    \item  AST detects some logical anomalies, but lacks an approach that detects logical anomalies by design.
For example, it detects that the additional blue cable connecting two splicing connectors in \Cref{fig:qualitative_1} causes unseen features.
Yet, the missing pushpin in the box of pushpins in \Cref{fig:qualitative_1} is also an unseen feature and does not cause a response in the anomaly map of AST.
This highlights the importance of a reliable approach to logical anomalies.
Furthermore, it shows the dependence of anomaly detection methods on the choice of the feature extractor.
As shown in \Cref{tab:distillation_backbones}, \ourmethod\ is robust to this choice.
\item DSR produces very precise segmentations, but also suffers from false positives, for example on the grid and the wood image of MVTec AD in \Cref{fig:qualitative_2}.
At times, it furthermore shows no response at all to defects.
\item FastFlow's anomaly maps contain a large amount of noise, i.e. false positive detections. This hinders the interpretability of its detection results.
\item GCAD succeeds at detecting logical anomalies, but has difficulty with some structural anomalies that other methods detect reliably, such as the scratches on the metal nut in \Cref{fig:qualitative_2} or the green capsules in \Cref{fig:qualitative_3}.
\item \patchcoreens\ struggles with very small defects such as those of the printed circuit boards in \Cref{fig:qualitative_3}.
Small defects are challenging, but highly relevant for practical applications.
A small contamination can cause a high economic damage if it goes unnoticed, for example, in a pharmaceutical application.
\item SimpleNet performs similar to other methods on MVTec AD, but struggles with the more challenging anomalies of MVTec LOCO and VisA, for example the defective capsules and PCBs in \Cref{fig:qualitative_3}.
\item S--T is a patch-based anomaly detection approach and therefore can only detect anomalies if they involve patches that are anomalous per se, i.e., without putting them in the global context of the respective image.
While AST's feature vectors have a receptive field that spans across the entire image, S--T's receptive field is limited to 65$\times$65 pixels.
Therefore, it does not detect anomalies such as the missing transistor in \Cref{fig:qualitative_2}.
\end{itemize}

The qualitative results show tendencies of each method regarding the behavior on anomalous images.
While these results are informative, they should not be used exclusively for evaluating the anomaly detection performance of a method or for comparing methods.
\textbf{For that, metrics such as the AU-ROC and the AU-PRO are well-suited, since they are evaluated objectively on thousands of test images across dataset collections.}

{\small
\begin{thebibliography}{10}\itemsep=-1pt

\bibitem{akcay2022anomalib}
Samet Akcay, Dick Ameln, Ashwin Vaidya, Barath Lakshmanan, Nilesh Ahuja, and
  Utku Genc.
\newblock Anomalib: A deep learning library for anomaly detection.
\newblock In {\em 2022 IEEE International Conference on Image Processing
  (ICIP)}, pages 1706--1710. IEEE, 2022.

\bibitem{akcay2019ganomaly}
Samet Akcay, Amir Atapour-Abarghouei, and Toby~P Breckon.
\newblock Ganomaly: Semi-supervised anomaly detection via adversarial training.
\newblock In {\em Computer Vision--ACCV 2018: 14th Asian Conference on Computer
  Vision, Perth, Australia, December 2--6, 2018, Revised Selected Papers, Part
  III 14}, pages 622--637. Springer, 2019.

\bibitem{armato2011lung}
Samuel~G Armato~III, Geoffrey McLennan, Luc Bidaut, Michael~F McNitt-Gray,
  Charles~R Meyer, Anthony~P Reeves, Binsheng Zhao, Denise~R Aberle, Claudia~I
  Henschke, Eric~A Hoffman, et~al.
\newblock The lung image database consortium (lidc) and image database resource
  initiative (idri): a completed reference database of lung nodules on ct
  scans.
\newblock {\em Medical physics}, 38(2):915--931, 2011.

\bibitem{bailey2012machine_vision_handbook_fpgas}
Donald Bailey.
\newblock {\em Implementing Machine Vision Systems Using FPGAs}, pages
  1103--1136 in ``Machine Vision Handbook'' by Bruce G. Batchelor.
\newblock Springer London, London, 2012.

\bibitem{bakas2017_brats_dataset}
Spyridon Bakas, Hamed Akbari, Aristeidis Sotiras, Michel Bilello, Martin
  Rozycki, Justin~S. Kirby, et~al.
\newblock Advancing the cancer genome atlas glioma {MRI} collections with
  expert segmentation labels and radiomic features.
\newblock {\em Scientific Data}, 4(1), 2017.

\bibitem{baur2019_ano_vae_gan}
Christoph Baur, Benedikt Wiestler, Shadi Albarqouni, and Nassir Navab.
\newblock Deep autoencoding models for unsupervised anomaly segmentation in
  brain mr images.
\newblock In {\em Brainlesion: Glioma, Multiple Sclerosis, Stroke and Traumatic
  Brain Injuries}, pages 161--169. Springer International Publishing, 2019.

\bibitem{bergmann2021_mvtec_ad_ijcv}
Paul Bergmann, Kilian Batzner, Michael Fauser, David Sattlegger, and Carsten
  Steger.
\newblock {The MVTec Anomaly Detection Dataset: A Comprehensive Real-World
  Dataset for Unsupervised Anomaly Detection}.
\newblock {\em International Journal of Computer Vision}, 129(4):1038--1059,
  2021.

\bibitem{bergmann2021_mvtec_loco_ijcv}
Paul Bergmann, Kilian Batzner, Michael Fauser, David Sattlegger, and Carsten
  Steger.
\newblock {Beyond Dents and Scratches: Logical Constraints in Unsupervised
  Anomaly Detection and Localization}.
\newblock {\em International Journal of Computer Vision}, 130(4):947–--969,
  2022.

\bibitem{bergmann2019_mvtec_ad_cvpr}
Paul Bergmann, Michael Fauser, David Sattlegger, and Carsten Steger.
\newblock {MVTec AD --- A Comprehensive Real-World Dataset for Unsupervised
  Anomaly Detection}.
\newblock In {\em IEEE Conference on Computer Vision and Pattern Recognition
  (CVPR)}, pages 9584--9592, 2019.

\bibitem{bergmann2020_uninformed_cvpr}
Paul Bergmann, Michael Fauser, David Sattlegger, and Carsten Steger.
\newblock {Uninformed Students: Student-Teacher Anomaly Detection With
  Discriminative Latent Embeddings}.
\newblock In {\em IEEE Conference on Computer Vision and Pattern Recognition
  (CVPR)}, pages 4182--4191, 2020.

\bibitem{bergmann2022_mvtec_3dad}
Paul Bergmann, Xin Jin, David Sattlegger, and Carsten Steger.
\newblock {The MVTec 3D-AD Dataset for Unsupervised 3D Anomaly Detection and
  Localization}.
\newblock In {\em Proceedings of the 17th International Joint Conference on
  Computer Vision, Imaging and Computer Graphics Theory and Applications -
  Volume 5: VISAPP}, pages 202--213. INSTICC, SciTePress, 2022.

\bibitem{bergmann2018_ssim_ae}
Paul Bergmann, Sindy Löwe, Michael Fauser, David Sattlegger, and Carsten
  Steger.
\newblock {Improving Unsupervised Defect Segmentation by Applying Structural
  Similarity to Autoencoders}.
\newblock In {\em Proceedings of the 14th International Joint Conference on
  Computer Vision, Imaging and Computer Graphics Theory and Applications -
  Volume 5: VISAPP}, pages 372--380. INSTICC, SciTePress, 2019.

\bibitem{blum2019_fishyscapes_dataset}
Hermann Blum, Paul-Edouard Sarlin, Juan Nieto, Roland Siegwart, and Cesar
  Cadena.
\newblock {Fishyscapes: A Benchmark for Safe Semantic Segmentation in
  Autonomous Driving}.
\newblock In {\em IEEE International Conference on Computer Vision Workshops
  (ICCVW)}, pages 2403--2412, 2019.

\bibitem{cohen2020_subimage}
Niv Cohen and Yedid Hoshen.
\newblock Sub-image anomaly detection with deep pyramid correspondences.
\newblock {\em arXiv preprint arXiv:2005.02357v1}, 2020.

\bibitem{davis2006relationship}
Jesse Davis and Mark Goadrich.
\newblock The relationship between precision-recall and roc curves.
\newblock In {\em Proceedings of the 23rd international conference on Machine
  learning}, pages 233--240, 2006.

\bibitem{defard2021_PaDiM}
Thomas Defard, Aleksandr Setkov, Angelique Loesch, and Romaric Audigier.
\newblock {PaDiM: A Patch Distribution Modeling Framework for Anomaly Detection
  and Localization}.
\newblock In {\em Pattern Recognition. ICPR International Workshops and
  Challenges}, pages 475--489. Springer International Publishing, 2021.

\bibitem{brox2016_learned_visual_similarity_metrics}
Alexey Dosovitskiy and Thomas Brox.
\newblock {Generating Images with Perceptual Similarity Metrics based on Deep
  Networks}.
\newblock In {\em Advances in Neural Information Processing Systems}, pages
  658--666, 2016.

\bibitem{ehret2019_ad_review_paper}
Thibaud Ehret, Axel Davy, Jean-Michel Morel, and Mauricio Delbracio.
\newblock {Image Anomalies: A Review and Synthesis of Detection Methods}.
\newblock {\em Journal of Mathematical Imaging and Vision}, 61(5):710--743,
  2019.

\bibitem{fernando2021_medical_ad_survey}
Tharindu Fernando, Harshala Gammulle, Simon Denman, Sridha Sridharan, and
  Clinton Fookes.
\newblock Deep learning for medical anomaly detection – a survey.
\newblock {\em ACM Computing Surveys}, 54(7), 2021.

\bibitem{gong2019_mem_ae_iccv}
Dong Gong, Lingqiao Liu, Vuong Le, Budhaditya Saha, Moussa~Reda Mansour, Svetha
  Venkatesh, and Anton Van Den~Hengel.
\newblock Memorizing normality to detect anomaly: Memory-augmented deep
  autoencoder for unsupervised anomaly detection.
\newblock In {\em IEEE International Conference on Computer Vision (ICCV)},
  pages 1705--1714, 2019.

\bibitem{goodfellow2014_gans}
Ian Goodfellow, Jean Pouget-Abadie, Mehdi Mirza, Bing Xu, David Warde-Farley,
  Sherjil Ozair, Aaron Courville, and Yoshua Bengio.
\newblock {Generative Adversarial Nets}.
\newblock In {\em Advances in Neural Information Processing Systems}, pages
  2672--2680, 2014.

\bibitem{gudovskiy2022_CFLOW}
Denis Gudovskiy, Shun Ishizaka, and Kazuki Kozuka.
\newblock {CFLOW-AD: Real-Time Unsupervised Anomaly Detection With Localization
  via Conditional Normalizing Flows}.
\newblock In {\em Proceedings of the IEEE/CVF Winter Conference on Applications
  of Computer Vision (WACV)}, pages 98--107, 2022.

\bibitem{he2016_resnet_paper}
Kaiming He, Xiangyu Zhang, Shaoqing Ren, and Jian Sun.
\newblock {Deep Residual Learning for Image Recognition}.
\newblock In {\em IEEE Conference on Computer Vision and Pattern Recognition
  (CVPR)}, pages 770--778, 2016.

\bibitem{hendrycks2019scaling}
Dan Hendrycks, Steven Basart, Mantas Mazeika, Andy Zou, Joseph Kwon,
  Mohammadreza Mostajabi, Jacob Steinhardt, and Dawn Song.
\newblock Scaling out-of-distribution detection for real-world settings.
\newblock In Kamalika Chaudhuri, Stefanie Jegelka, Le Song, Csaba Szepesvari,
  Gang Niu, and Sivan Sabato, editors, {\em Proceedings of the 39th
  International Conference on Machine Learning}, volume 162 of {\em Proceedings
  of Machine Learning Research}, pages 8759--8773. PMLR, 17--23 Jul 2022.

\bibitem{huang2017densely}
Gao Huang, Zhuang Liu, Laurens Van Der~Maaten, and Kilian~Q Weinberger.
\newblock Densely connected convolutional networks.
\newblock In {\em Proceedings of the IEEE conference on computer vision and
  pattern recognition}, pages 4700--4708, 2017.

\bibitem{huang2018_magnetic_tile_dataset}
Yibin Huang, Congying Qiu, Yue Guo, Xiaonan Wang, and Kui Yuan.
\newblock Surface defect saliency of magnetic tile.
\newblock In {\em 2018 IEEE 14th International Conference on Automation Science
  and Engineering (CASE)}, pages 612--617, 2018.

\bibitem{irvin2019chexpert}
Jeremy Irvin, Pranav Rajpurkar, Michael Ko, Yifan Yu, Silviana Ciurea-Ilcus,
  Chris Chute, Henrik Marklund, Behzad Haghgoo, Robyn Ball, Katie Shpanskaya,
  et~al.
\newblock Chexpert: A large chest radiograph dataset with uncertainty labels
  and expert comparison.
\newblock In {\em Proceedings of the AAAI conference on artificial
  intelligence}, volume~33, pages 590--597, 2019.

\bibitem{jezek2021deep}
Stepan Jezek, Martin Jonak, Radim Burget, Pavel Dvorak, and Milos Skotak.
\newblock Deep learning-based defect detection of metal parts: evaluating
  current methods in complex conditions.
\newblock In {\em 2021 13th International Congress on Ultra Modern
  Telecommunications and Control Systems and Workshops (ICUMT)}, pages 66--71.
  IEEE, 2021.

\bibitem{kingma2014_adam}
Diederik~P Kingma and Jimmy Ba.
\newblock {Adam: A Method for Stochastic Optimization}.
\newblock {\em International Conference on Learning Representations (ICLR)},
  2015.

\bibitem{li2021cutpaste}
Chun-Liang Li, Kihyuk Sohn, Jinsung Yoon, and Tomas Pfister.
\newblock Cutpaste: Self-supervised learning for anomaly detection and
  localization.
\newblock In {\em Proceedings of the IEEE/CVF Conference on Computer Vision and
  Pattern Recognition}, pages 9664--9674, 2021.

\bibitem{li2013_ucsd_video_ad_dataset}
Wei-Xin Li, Vijay Mahadevan, and Nuno Vasconcelos.
\newblock {Anomaly Detection and Localization in Crowded Scenes}.
\newblock {\em IEEE Transactions on Pattern Analysis and Machine Intelligence
  (TPAMI)}, 36(1):18--32, 2013.

\bibitem{lis2019_iccv_resynthesis}
Krzysztof Lis, Krishna~Kanth Nakka, Pascal Fua, and Mathieu Salzmann.
\newblock Detecting the unexpected via image resynthesis.
\newblock In {\em IEEE International Conference on Computer Vision (ICCV)},
  pages 2152--2161, 2019.

\bibitem{liu2020_visually_explaining_vaes}
Wenqian Liu, Runze Li, Meng Zheng, Srikrishna Karanam, Ziyan Wu, Bir Bhanu,
  Richard~J. Radke, and Octavia Camps.
\newblock Towards visually explaining variational autoencoders.
\newblock In {\em IEEE Conference on Computer Vision and Pattern Recognition
  (CVPR)}, pages 8639--8648, 2020.

\bibitem{liu2021swin}
Ze Liu, Yutong Lin, Yue Cao, Han Hu, Yixuan Wei, Zheng Zhang, Stephen Lin, and
  Baining Guo.
\newblock Swin transformer: Hierarchical vision transformer using shifted
  windows.
\newblock In {\em Proceedings of the IEEE/CVF international conference on
  computer vision}, pages 10012--10022, 2021.

\bibitem{liu2022convnet}
Zhuang Liu, Hanzi Mao, Chao-Yuan Wu, Christoph Feichtenhofer, Trevor Darrell,
  and Saining Xie.
\newblock A convnet for the 2020s.
\newblock In {\em Proceedings of the IEEE/CVF Conference on Computer Vision and
  Pattern Recognition}, pages 11976--11986, 2022.

\bibitem{Liu_2023_CVPR}
Zhikang Liu, Yiming Zhou, Yuansheng Xu, and Zilei Wang.
\newblock {Simplenet: A simple network for image anomaly detection and localization}.
\newblock In {\em Proceedings of the IEEE/CVF Conference on Computer Vision and Pattern Recognition (CVPR)}, pages 20402--20411, June 2023.

\bibitem{lu2013_avenue_video_ad_dataset}
Cewu Lu, Jianping Shi, and Jiaya Jia.
\newblock {Abnormal Event Detection at 150 FPS in MATLAB}.
\newblock In {\em IEEE International Conference on Computer Vision (ICCV)},
  pages 2720--2727, 2013.

\bibitem{menze2015_brats_dataset}
Bjoern~H. Menze, Andras Jakab, Stefan Bauer, Jayashree Kalpathy-Cramer, Keyvan
  Farahani, Justin Kirby, et~al.
\newblock {The Multimodal Brain Tumor Image Segmentation Benchmark ({BRATS})}.
\newblock {\em IEEE Transactions on Medical Imaging}, 34(10):1993--2024, 2015.

\bibitem{mishra2021vt}
Pankaj Mishra, Riccardo Verk, Daniele Fornasier, Claudio Piciarelli, and
  Gian~Luca Foresti.
\newblock Vt-adl: A vision transformer network for image anomaly detection and
  localization.
\newblock In {\em 2021 IEEE 30th International Symposium on Industrial
  Electronics (ISIE)}, pages 01--06. IEEE, 2021.

\bibitem{napoletano2018_cnn_feature_dictionary_nanofibres}
Paolo Napoletano, Flavio Piccoli, and Raimondo Schettini.
\newblock {Anomaly Detection in Nanofibrous Materials by {CNN}-Based
  Self-Similarity}.
\newblock {\em Sensors}, 18(1):209, 2018.

\bibitem{nazare2018_pretrained_cnns_for_ad}
Tiago~S Nazare, Rodrigo~F de Mello, and Moacir~A Ponti.
\newblock Are pre-trained cnns good feature extractors for anomaly detection in
  surveillance videos?
\newblock {\em arXiv preprint arXiv:1811.08495v1}, 2018.

\bibitem{odena2016deconvolution}
Augustus Odena, Vincent Dumoulin, and Chris Olah.
\newblock Deconvolution and checkerboard artifacts.
\newblock {\em Distill}, 2016.

\bibitem{pang2020_review_paper}
Guansong Pang, Chunhua Shen, Longbing Cao, and Anton Van~Den Hengel.
\newblock Deep learning for anomaly detection: A review.
\newblock {\em ACM Comput. Surv.}, 54(2), mar 2021.

\bibitem{park2020_mmnad}
Hyunjong Park, Jongyoun Noh, and Bumsub Ham.
\newblock Learning memory-guided normality for anomaly detection.
\newblock In {\em IEEE Conference on Computer Vision and Pattern Recognition
  (CVPR)}, pages 14360--14369, 2020.

\bibitem{paszke2019_PyTorch}
Adam Paszke, Sam Gross, Francisco Massa, Adam Lerer, James Bradbury, Gregory
  Chanan, Trevor Killeen, Zeming Lin, Natalia Gimelshein, Luca Antiga, Alban
  Desmaison, Andreas Kopf, Edward Yang, Zachary DeVito, Martin Raison, Alykhan
  Tejani, Sasank Chilamkurthy, Benoit Steiner, Lu Fang, Junjie Bai, and Soumith
  Chintala.
\newblock {PyTorch: An Imperative Style, High-Performance Deep Learning
  Library}.
\newblock In {\em {Advances in Neural Information Processing Systems}},
  volume~32, 2019.

\bibitem{perera2019_cvpr_ocgan}
Pramuditha Perera, Ramesh Nallapati, and Bing Xiang.
\newblock Ocgan: One-class novelty detection using gans with constrained latent
  representations.
\newblock In {\em IEEE Conference on Computer Vision and Pattern Recognition
  (CVPR)}, pages 2893--2901, 2019.

\bibitem{redmon2016you}
Joseph Redmon, Santosh Divvala, Ross Girshick, and Ali Farhadi.
\newblock You only look once: Unified, real-time object detection.
\newblock In {\em Proceedings of the IEEE conference on computer vision and
  pattern recognition}, pages 779--788, 2016.

\bibitem{rezende2015variational}
Danilo Rezende and Shakir Mohamed.
\newblock Variational inference with normalizing flows.
\newblock In {\em International conference on machine learning}, pages
  1530--1538. PMLR, 2015.

\bibitem{rippel2021_Gaussian}
Oliver Rippel., Arnav Chavan., Chucai Lei., and Dorit Merhof.
\newblock Transfer learning gaussian anomaly detection by fine-tuning
  representations.
\newblock In {\em Proceedings of the 2nd International Conference on Image
  Processing and Vision Engineering - IMPROVE,}, pages 45--56. INSTICC,
  SciTePress, 2022.

\bibitem{ronneberger2015_u_net}
Olaf Ronneberger, Philipp Fischer, and Thomas Brox.
\newblock U-net: Convolutional networks for biomedical image segmentation.
\newblock In {\em Medical Image Computing and Computer-Assisted Intervention --
  MICCAI 2015}, pages 234--241. Springer International Publishing, 2015.

\bibitem{roth2022towards}
Karsten Roth, Latha Pemula, Joaquin Zepeda, Bernhard Sch{\"o}lkopf, Thomas
  Brox, and Peter Gehler.
\newblock Towards total recall in industrial anomaly detection.
\newblock In {\em Proceedings of the IEEE/CVF Conference on Computer Vision and
  Pattern Recognition}, pages 14318--14328, 2022.

\bibitem{rudolph2021_differnet}
Marco Rudolph, Bastian Wandt, and Bodo Rosenhahn.
\newblock Same same but differnet: Semi-supervised defect detection with
  normalizing flows.
\newblock In {\em 2021 IEEE Winter Conference on Applications of Computer
  Vision (WACV)}, pages 1906--1915, 2021.

\bibitem{rudolph2022_cross_flows}
Marco Rudolph, Tom Wehrbein, Bodo Rosenhahn, and Bastian Wandt.
\newblock Fully convolutional cross-scale-flows for image-based defect
  detection.
\newblock In {\em 2022 IEEE/CVF Winter Conference on Applications of Computer
  Vision (WACV)}, pages 1829--1838, 2022.

\bibitem{rudolph2023asymmetric}
Marco Rudolph, Tom Wehrbein, Bodo Rosenhahn, and Bastian Wandt.
\newblock Asymmetric student-teacher networks for industrial anomaly detection.
\newblock In {\em Proceedings of the IEEE/CVF Winter Conference on Applications
  of Computer Vision}, pages 2592--2602, 2023.

\bibitem{russakovsky2015_alexnet}
Olga Russakovsky, Jia Deng, Hao Su, Jonathan Krause, Sanjeev Satheesh, Sean Ma,
  Zhiheng Huang, Andrej Karpathy, Aditya Khosla, Michael Bernstein,
  Alexander~C. Berg, and Li Fei-Fei.
\newblock Imagenet large scale visual recognition challenge.
\newblock {\em International Journal of Computer Vision}, 115(3):211--252,
  2015.

\bibitem{sakurada2014_aes_for_ad}
Mayu Sakurada and Takehisa Yairi.
\newblock Anomaly detection using autoencoders with nonlinear dimensionality
  reduction.
\newblock In {\em Proceedings of the MLSDA 2014 2nd Workshop on Machine
  Learning for Sensory Data Analysis}, MLSDA'14, page 4–11, New York, NY,
  USA, 2014. Association for Computing Machinery.

\bibitem{salehi2021_st_ad}
Mohammadreza Salehi, Niousha Sadjadi, Soroosh Baselizadeh, Mohammad~H. Rohban,
  and Hamid~R. Rabiee.
\newblock Multiresolution knowledge distillation for anomaly detection.
\newblock In {\em IEEE Conference on Computer Vision and Pattern Recognition
  (CVPR)}, pages 14897--14907, 2021.

\bibitem{sandler2018mobilenetv2}
Mark Sandler, Andrew Howard, Menglong Zhu, Andrey Zhmoginov, and Liang-Chieh
  Chen.
\newblock Mobilenetv2: Inverted residuals and linear bottlenecks.
\newblock In {\em Proceedings of the IEEE conference on computer vision and
  pattern recognition}, pages 4510--4520, 2018.

\bibitem{schlegl2017_anogan}
Thomas Schlegl, Philipp Seeb{\"o}ck, Sebastian~M Waldstein, Ursula
  Schmidt-Erfurth, and Georg Langs.
\newblock {Unsupervised Anomaly Detection with Generative Adversarial Networks
  to Guide Marker Discovery}.
\newblock In {\em International Conference on Information Processing in Medical
  Imaging}, pages 146--157. Springer, 2017.

\bibitem{schlegl2019_fast_anogan}
Thomas Schlegl, Philipp Seeböck, Sebastian Waldstein, Georg Langs, and Ursula
  Schmidt-Erfurth.
\newblock {f-AnoGAN: Fast Unsupervised Anomaly Detection with Generative
  Adversarial Networks}.
\newblock {\em Medical Image Analysis}, 54, 2019.

\bibitem{shrivastava2016training}
Abhinav Shrivastava, Abhinav Gupta, and Ross Girshick.
\newblock Training region-based object detectors with online hard example
  mining.
\newblock In {\em Proceedings of the IEEE conference on computer vision and
  pattern recognition}, pages 761--769, 2016.

\bibitem{steger2018_mva_book}
Carsten Steger, Markus Ulrich, and Christian Wiedemann.
\newblock {\em Machine Vision Algorithms and Applications}.
\newblock Wiley-VCH, Weinheim, 2nd edition, 2018.

\bibitem{tan2019efficientnet}
Mingxing Tan and Quoc Le.
\newblock Efficientnet: Rethinking model scaling for convolutional neural
  networks.
\newblock In {\em International conference on machine learning}, pages
  6105--6114. PMLR, 2019.

\bibitem{tan2021efficientnetv2}
Mingxing Tan and Quoc Le.
\newblock Efficientnetv2: Smaller models and faster training.
\newblock In {\em International conference on machine learning}, pages
  10096--10106. PMLR, 2021.

\bibitem{tan2020efficientdet}
Mingxing Tan, Ruoming Pang, and Quoc~V Le.
\newblock Efficientdet: Scalable and efficient object detection.
\newblock In {\em Proceedings of the IEEE/CVF conference on computer vision and
  pattern recognition}, pages 10781--10790, 2020.

\bibitem{wang2021student_teacher}
Guodong Wang, Shumin Han, Errui Ding, and Di Huang.
\newblock Student-teacher feature pyramid matching for anomaly detection.
\newblock In {\em The British Machine Vision Conference (BMVC)}, 2021.

\bibitem{xie2017resnext}
Saining Xie, Ross Girshick, Piotr Doll{\'a}r, Zhuowen Tu, and Kaiming He.
\newblock Aggregated residual transformations for deep neural networks.
\newblock In {\em Proceedings of the IEEE conference on computer vision and
  pattern recognition}, pages 1492--1500, 2017.

\bibitem{yin2022vit}
Hongxu Yin, Arash Vahdat, Jose~M Alvarez, Arun Mallya, Jan Kautz, and Pavlo
  Molchanov.
\newblock A-vit: Adaptive tokens for efficient vision transformer.
\newblock In {\em Proceedings of the IEEE/CVF Conference on Computer Vision and
  Pattern Recognition}, pages 10809--10818, 2022.

\bibitem{yu2021fastflow}
Jiawei Yu, Ye Zheng, Xiang Wang, Wei Li, Yushuang Wu, Rui Zhao, and Liwei Wu.
\newblock Fastflow: Unsupervised anomaly detection and localization via 2d
  normalizing flows.
\newblock {\em arXiv preprint arXiv:2111.07677v1}, 2021.

\bibitem{zagoruyko2016wideresnet_wrn}
Sergey Zagoruyko and Nikos Komodakis.
\newblock Wide residual networks.
\newblock In Edwin R.~Hancock Richard C.~Wilson and William A.~P. Smith,
  editors, {\em Proceedings of the British Machine Vision Conference (BMVC)},
  pages 87.1--87.12. BMVA Press, September 2016.

\bibitem{zavrtanik2022dsr}
Vitjan Zavrtanik, Matej Kristan, and Danijel Sko{\v{c}}aj.
\newblock Dsr--a dual subspace re-projection network for surface anomaly
  detection.
\newblock In {\em Computer Vision--ECCV 2022: 17th European Conference, Tel
  Aviv, Israel, October 23--27, 2022, Proceedings, Part XXXI}, pages 539--554.
  Springer, 2022.

\bibitem{zhang2016single}
Yingying Zhang, Desen Zhou, Siqin Chen, Shenghua Gao, and Yi Ma.
\newblock Single-image crowd counting via multi-column convolutional neural
  network.
\newblock In {\em Proceedings of the IEEE conference on computer vision and
  pattern recognition}, pages 589--597, 2016.

\bibitem{zong2018deep}
Bo Zong, Qi Song, Martin~Renqiang Min, Wei Cheng, Cristian Lumezanu, Daeki Cho,
  and Haifeng Chen.
\newblock Deep autoencoding gaussian mixture model for unsupervised anomaly
  detection.
\newblock In {\em International conference on learning representations}, 2018.

\bibitem{zou2022spot}
Yang Zou, Jongheon Jeong, Latha Pemula, Dongqing Zhang, and Onkar Dabeer.
\newblock Spot-the-difference self-supervised pre-training for anomaly
  detection and segmentation.
\newblock In {\em Computer Vision--ECCV 2022: 17th European Conference, Tel
  Aviv, Israel, October 23--27, 2022, Proceedings, Part XXX}, pages 392--408.
  Springer, 2022.

\end{thebibliography}

}
\end{document}